\documentclass[Proceedings]{ascelike}
\usepackage{graphicx}
\usepackage{subfigure}
\usepackage{amsmath}
\usepackage{amsfonts}
\usepackage{amssymb}
\usepackage{amsbsy}
\usepackage{natbib}
\DeclareMathOperator*{\argmax}{argmax}

\usepackage{algorithm}
\usepackage{algorithmic}
\usepackage{multirow}
\usepackage{booktabs} 
\NameTag{Manuscript, A journal}
\begin{document}
%
\title{Differential Variable Speed Limits Control for Freeway Recurrent Bottlenecks via Deep Reinforcement learning}
\author{
Yuankai Wu%
%
\thanks{
School of Mechanical Engineering.,
Beijing Institute of Technology., 
No.5 Yard, Zhong Guan Cun South Street Haidian District.,
Beijing, OR 100081. E-mail: Kaimaogege@gmail.com.},
\ Huachun Tan\thanks{
School of Transportation Engineering.,
Southeast University., 
Sipailou 2, Xuanwu District.,
Nanjing, OR 210096. E-mail: tanhc@seu.edu.cn.},
\ Bin Ran\thanks{
Civil and Environmental Engineering.,
University of Wisconsin-Madison.,
2312 Engineering Hall, 1415 Engineering Drive., 
Madison, WI, OR 53706, E-mail: bran@engr.wisc.edu.
}
%
%
%
%
%
%
}
\maketitle
\begin{abstract}
Variable speed limits (VSL) control is a flexible way to improve traffic condition, increase safety and reduce emission. There is an emerging trend of using reinforcement learning technique for VSL control and recent studies have shown promising results. Currently, deep learning is enabling reinforcement learning to develope autonomous control agents for problems that were previously intractable. In this paper, we propose a more effective deep reinforcement learning (DRL) model for differential variable speed limits (DVSL) control, in which the dynamic and different speed limits among lanes can be imposed. The proposed DRL models use a novel actor-critic architecture which can learn a large number of discrete speed limits in a continues action space. Different reward signals, e.g. total travel time, bottleneck speed, emergency braking, and vehicular emission are used to train the DVSL controller, and comparison between these reward signals are conducted. We test proposed DRL baased DVSL controllers on a simulated freeway recurrent bottleneck. Results show that the efficiency, safety and emissions can be improved by the proposed method. We also show some interesting findings through the visulization of the control policies generated from DRL models.
\end{abstract}
%
%
\section{Introduction}
\label{In}
Variable speed limits (VSL) or speed harmonization has been studied for a long history \citep{khondaker2015variable,lu2014review}. Specially, VSL have been verified to improve traffic safety, resolve traffic breakdown and bring environmental benefits. For example, the applications of VSL in German had shown that VSL typically results in lower accident rates and $5\% \sim 10\%$ increase in capacity \citep{weikl2013traffic}. In English case, the VSL increased $7\%$ capacity and decreased the overall congestion time \citep{middelham2006dynamic}. Neitherlands has also applied VSL, $20\% \sim 30\%$ traffic emission reductions of NOx and PM10 were reported in test locations \citep{macdonald2008atm}. VSL control is thus regarded as a hot topic in intelligent transportation systems.

Prior studies in VSL control can be categorized into: (\romannumeral1) hand-crafted rules based strategies \citep{soriguera2013assessment,piao2008safety}, in which the speed limits are controlled with pre-defined rules, the VSL decisions are changed with pre-defined time plan or pre-selected thresholds of traffic state. (\romannumeral2) proactive approaches \citep{hegyi2005optimal,kattan2015probe}, in which future traffic is predicted so that traffic bottlenecks are anticipated, the controller automatically adjusts the speed limits to avoid traffic breakdowns. Previous studies have some limits. For example, the rule based approaches require expert knowledge from experienced engineers. In addition, its robustness and generality cannot be guaranteed. The performance of proactive approaches are heavily dependent on the
accuracy of traffic state prediction algorithm. More specifically, the VSL controller takes
actions shortly after a prediction time lag. If traffic state is largely varying over time, the
controller may not be able to achieve the desired performance in real-time.

\begin{figure}[h!] 
\begin{center}
\includegraphics[width=0.7\textwidth]{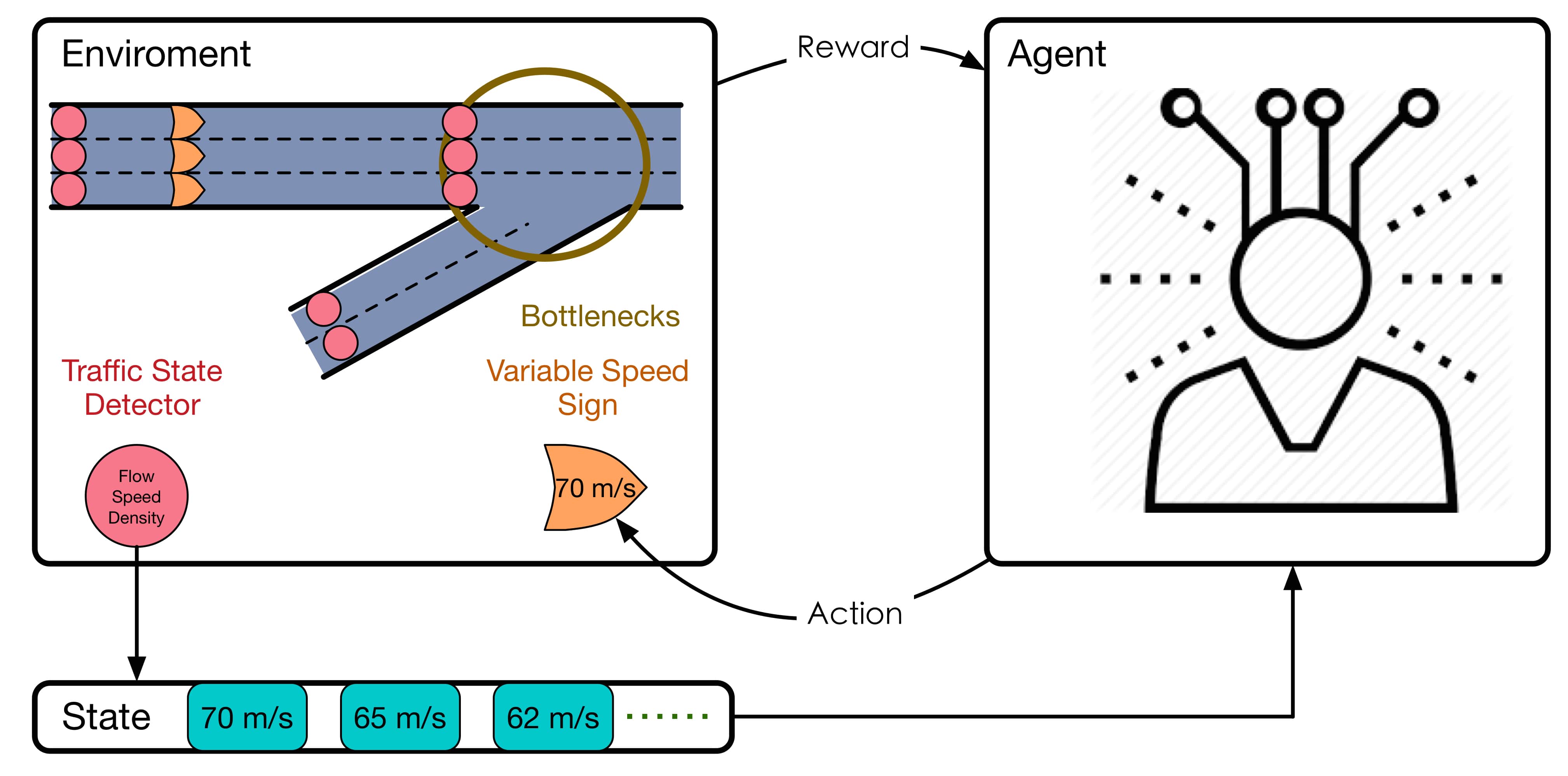}
\caption{Reinforcement learning framework for variable speed limit control.}
\label{VSL_Framwork}
\end{center}
\end{figure}

Recently, the emergence of reinforcement learning provides great potential for addressing the limitations associated with state-of-art VSL control strategies \citep{li2017reinforcement,zhu2014accounting}. The reinforcement learning (RL) studied how artificial agents take actions in an environment so as to maximize its long-term cumulative reward. As a result, a well-trained RL agent can theoretically achieve a proactive control scheme to optimize its benefits. The framework of RL based VSL control is given in Fig. \ref{VSL_Framwork}. For RL based VSL control, environment is composed of a transportation networks with bottlenecks, traffic state detectors and the variable speed signs. State is a feature representation of the traffic state collected from detectors. Agent takes state as input and learns a model to change the speed limits. The speed limits are sent to the variable speed signs of enviroment and the reward (e.g., the traffic state of bottlenecks) is sent back to the agent.  

Two unique challenges arise in applying traditional RL to VSL control: (\romannumeral1) The difficulties in state summarization and representation. (\romannumeral2) The limited capability for learning the correlation between enviroment and optimal speed limits. In related fields including ramp metering \citep{belletti2018expert} and traffic light control \citep{van2016coordinated,wei2018intellilight}, recent studies have applied deep reinforcement learning (DRL) techniques to address these two challenges. In DRL framework, massive traffic information collected from a large amount of traffic detectors can be described as a vector or an image and is directly taken as an input for neural networks of DRL. By the powerful function approximation properties of neural networks, the DRL agent can learn optimal phase of traffic lights, succeeding controllers that used to be hand-engineered. 

Another motivation of this study is to handle the VSL control with the emergence of connected and autonomous vehicles (CAV) technologies \citep{roncoli2015traffic}. Though CAV has not been implemented in the real world transportation system yet, it is believed that the CAV technologies will improve the safety of traffic, alleviate congestion and reduce vehicular emission. In the authors' opinion, the CAV technologies will bring two major benefits to VSL control. (\romannumeral1) CAV facilitates vehicles to communicate with infrastructures such as traffic state detectors, traffic lights and variable speed signs. (\romannumeral2) CAV will ultimately lead to more rational and safe driving behavior, since bad drivers will be replaced by artificial intelligence. Mordern VSL control strategies such as differential variable speed limit (DVSL) among lanes is dramatically limited by driver compliance \citep{schick2003effects}. The impacts of VSL, in terms of safety and travel time, are quite sensitive to the level of driver compliance \citep{hellinga2011impact}. In CAV enviroment, the dynamical speed limits can be sent to each individuel autonomous vehicle, and the autonomous vehicles can be enforced to drive under the limits. For a sufficient penetration of autonomous vehicles, this will be sufficient to impose the speed limit to normal vehicles as well. As a result, the problem of driver compliance will be readily solved under a highly developed CAV eneviroment. 

In the light of advances in DRL and CAV techniques, in this paper we propose a novel DRL approach to learn highly efficient VSL control strategies for freeway recurrent bottlenecks. Our approach makes several important contributions: 

\textbf{1.A novel VSL control framework allows different and dynamic speed limits among lanes}: Differential VSL (DVSL) may be beneficial for numerous systems and reducing traffic congestion. However, currently the applications of DVSL have not been fully studied due to the issue of driver conpliance and implementation. Under the CAV enviroment, vehicles are able to communicate with speed limit controllers, the AI driver can be enforced to drive under the given speed limits. Hence DVSL will become technologically possible. In this paper, we evaluate the proposed method on a 5-lane freeway with a recurrent bottlenecks between on and off ramps. The VSL controller is used to dynamically set different speed limits for 5 lanes to reduce congestion.

\textbf{2. A DRL model for DVSL}: There is a significant technical challenges when modeling DVSL control using DRL. The DVSL contoller need to reason with a large number of possible actions at every step. For example, the total number of actions for a 5-lane freeway section with 10 speed limits will be $10^5 = 100000$. Though exsiting studies have shown that VSL control is fundamentally reinforcement problems, the traidional Q learning approach, deep Q networks and continues DRL framework are difficult or impossible to handle such a large action space. Shed light on the work \citep{dulac2015deep} of DeepMind, in this paper we present a deep policy architecture which operates efficiently with a large number of discrete speed limits. The policy first produces a continuous action, and then finds the set of closest speed limits. The policy architecture builds upon the actor-critic framework. The actor is used to generate speed limits, and the critic is to evaluate the speed limits generated from the actor. Both the actor and critic \citep{sutton2017reinforcement} are approximated as neural networks. 

\textbf{3. Comparisons between different reward signals}: In essence, the RL problem is reward-driving, meaning that the best actor is determined by the rewards provided by the environment \citep{arulkumaran2017brief}. The definition of reward is important for RL. The motivation of implementing VSL is various. A sucessful VSL strategy should be able to alleviate congestion, improve safety and reduce emission. The performance of DRL based VSL control is highly related to the design of reward signal. The reward could worsen some aspects of the VSL controller in some cases. For example, a reward signal motivating the agent to reduce congestion might lead to severe safety issues and increase the emissions. Therefore it is very meaningful to study the reward design for DRL based VSL control. In this paper, we used numerous reward signals to train VSL controllers with a same neural networks structure and conduct comparisons between them.

The rest of paper is organized as follows. We first give a literature review on the related works in Sec \ref{Rw}. Then the problem statement is introduced in Sec \ref{Ps}. The methdology is described in Sec \ref{DRLDVSL}. The experimental results are shown in Sec \ref{Ex}. Finally, we conclude the paper in Sec \ref{Con}.

\section{Related works}
\label{Rw}
In this section, we firstly introduce conventional methods for VSL control, then introduce deep reinforcement learning and its application on related applications for intelligent transportation systems.
\subsection{Conventional variable speed limits control}
As mentioned in Sec \ref{In}, VSL is roughly conducted in rule based or proactive ways. Early VSL studies were mainly formulated as rule-based logic due to its relatively simpler problem settings. Dynamic speed limits are set according to pre-defined thresholds of traffic flow, occupancy or mean speed. For example, \citet{abdel2008dynamic} use speed difference between different sections as the threshold for changing speed limit and indicate that VSL can reduce crash probability. In \citep{papageorgiou2008effects}, VSL system based on flow–-speed threshold is investigated. It is suggested that VSL could be used in the interest of traffic safety rather than efficiency. The rule-based VSL systems have also achieved success in improving throughput and reducing travel time \citep{lin2004exploring,lyles2004field}. The rule-based method largely depends on the current traffic condition, without using the predictive information. 

Proactive approaches additionally considers the predictive information compared with rule based approaches. \citet*{hegyi2005model} demonstrate that model predictive control for coordination control variable speed limits and ramp metering led to $15\%$ travel time reduction. \citet*{carlson2010optimal} show that the traffic flow can be substantially improved via proactive VSL control. The above mentioned studies and their similar models \citep{kattan2015probe,hadiuzzaman2013cell} all use prediction model to predict future traffic condition. Hence, the success of proactive approaches are based on robustness and reliability of the short-term traffic prediction model in representing future traffic states. However, the accurate and reliable short-term traffic prediction \citep{wu2018hybrid,li2018travel} is not an easy task because the evolution of traffic state is related to many factors.

\subsection{Deep reinforcement learning}
The essence of RL is learning through interaction. In a typical RL framework, an autonomous agent, controlled by a machine learning algorithm, observes a state from its environment. The agent interacts with the environment by taking an action. Then the environment transition to a new state based on the current state and the chosen action. The objective of the agent is to maximize the accumulated rewards returned by the environment. Historiccaly, RL had some successes in some areas \citep{singh2002optimizing,ng2006autonomous}. However, the lacked scalablity and complexity issues have limited their application. 

The advent of deep learning has dramatically improved RL. Typically ``deep reinforcement learning'' (DRL) is defined as the utilization of deep learning algorithms within RL. DRL shows impressive successes on a wide range of tasks involving playing video games \citep{mnih2015human}, defeating a human world champion in Go \citep{silver2016mastering}, controlling robots \citep{levine2018learning} and indoor navigation \citep{zhu2017target}. There are numerous DRL approaches including deep Q networks (DQN) \citep{mnih2015human}, Evolutionary Strategy (ES) \citep{salimans2017evolution} and various policy gradient methods, such as TRPO \citep{schulman2015trust}, A3C \citep{mnih2016asynchronous}, DDPG \citep{lillicrap2015continuous}, and PPO \citep{schulman2017proximal}. Those algorithms hold great promise for learning to solve challenging control problems.

Advances in DRL and big traffic data lead to potential applications of DRL techniques to tackle challenging control problems in intelligent transportation systems. DRL has given promising results in ramp metering \citep{belletti2018expert}, traffic light control \citep{van2016coordinated,wei2018intellilight} and fleet management \citep{lin2018efficient}. These works have close connections to VSL in terms of problem setting.

\section{Problem statement}
\label{Ps}
In this section, we first provide the differential VSL (DVSL) control example considered in this paper, and then briefly discuss how this could be viewed as an MDP.
\subsection{Differential VSL among lanes}
Variable Speed Limit (VSL) control enables dynamic changing of posted speed limits in response to prevailing traffic. The objective of VSL control is to maximize transportation network performance including efficiency, safety and environmental friendliness. From the perspective of traffic flow theory, VSL can slow down traffic and change inflow to the bottleneck, therefore keep bottleneck traffic operating near its capacity state \citep{khondaker2015variable}.

\begin{figure}[h!] 
\begin{center}
\includegraphics[width=0.7\textwidth]{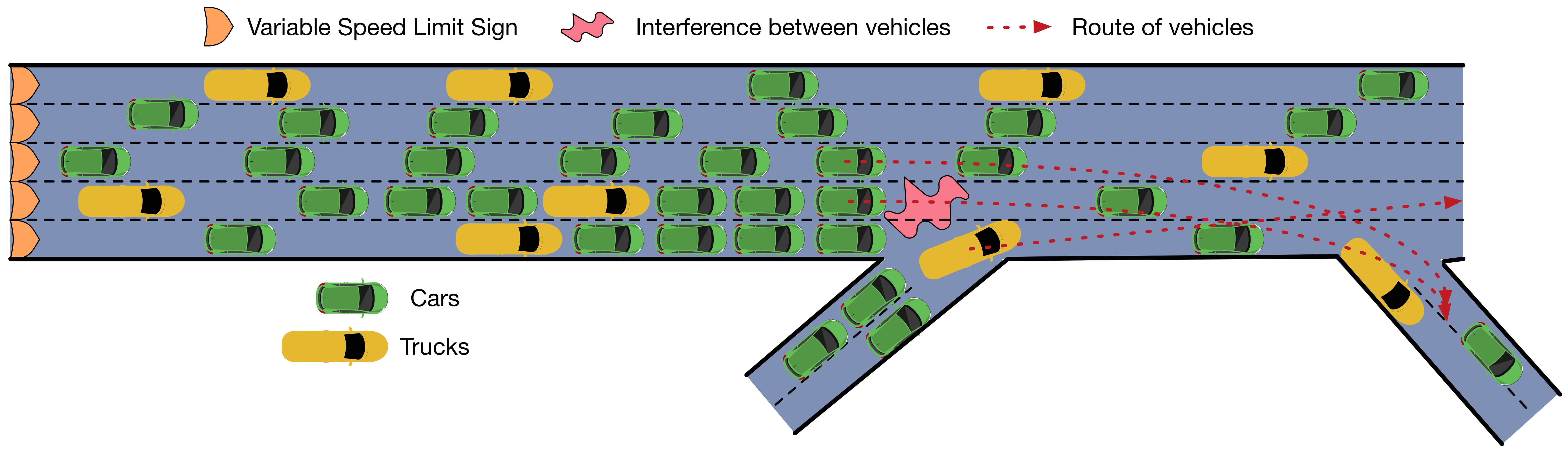}
\caption{The VSL control example considered in this paper.}
\label{DVSL_example}
\end{center}
\end{figure}

The VSL control example considered in this paper is illustrated in Fig. \ref{DVSL_example}. The freeway section in Fig. \ref{DVSL_example} is composed by five lanes and it presents an on-ramp and an off-ramp. As it may be seen in the figure, the interference between vehicles is appearing in the merge area between on-ramp and mainstream. The demand of exiting the freeway causes many car drivers to decide to move into the right lanes before the on-ramp. In this case, the interference between mainstream and on-ramp vehicles causes further speed reductions in the merging area, contributing to the creation of a generalised bottleneck. However, the left lanes of the freeway might be in the free-flow conditions when the bottleneck is formed. Traditionally, the same speed limit of all the 5 lanes are posted to control the outflow of the controlled section and prevente the capacity drop at active bottlenecks. In this paper we suppose that it is more efficient to post different speed limits among lanes. For example, there might be a short period that the merge area between on-ramp and mainstream is congested, whereas the left lanes are in free-flow conditions. In such cases, it is not necessary to reduce all the outflow of 5 lanes. In the contrary, a lower speed limit in left lanes would degrade the efficiency of the freeway. 

It should be noted that the DVSL among lanes has not been well researched until now. One reason may be that drivers are not familiar with the VSL system. In a case study of Germany, it is shown that the different speed limits motivate drivers to change into the fastest lanes and cause traffic breakdown and increase in crash probility \citep{schick2003effects}. The aim of this paper is to explore a mordern VSL solution under CAV enviroment. In CAV enviroment, it is not difficult to implement the DVSL control action by sending speed limit orders to the vehicles in the corresponding lane. Even the existing driver assistance systems e.g. fixed speed cruise control can be used to enforce the vehicle to drive under the received speed limit.

\subsection{Formulation as a markov decision process}
\label{MDPF}

To relate the DVSL control to the RL setting, the control problem should be formalized as a markov decision process (MDP), which involves trial-and-error interaction in an environment. An MDP consists of $(S, A, P, \mathcal{R}, \gamma)$, where $S$ denotes the state space, $P$ : $S \times A \times S$ is the transition probability, $R$ : $S \times A \times S \to M(R)$ is the reward distribution, and $\gamma \in [0, 1)$ is the discount factor. At each discrete time step $t = 1, 2, 3, \cdots$ the agent selects an action according to some policy $\pi$ and the environment responds by transitioning into a new state $s_{t+1}$ sampled from $p(:\mid s_t,a_t)$, and the agent receives a scalar reward $r_{t+1}$. The agent's goal is to maximize discounted cumulative sum of rewards, from the current state $s_t$, for some discount factor $\gamma < 1$. The formulations of VSL control as an MDP can be found in \citep{li2017reinforcement,zhu2014accounting}. In this paper, we extend the formalism to DVSL control. The definitions of agent, state, action, transition, and reward function are given as follows:

\textbf{Agent}: We consider a VSL controller as an agent. The agent can set different speed limits for each lane of its corresponding sections. For a transportation network or a freeway corridor, there will be a large number of availiable VSL controllers. In such case, the VSL control problem can be formulated as a multi-agent RL problem. In this paper, we only consider the DVSL control with a single agent. The goal of the agent is to improve the efficiency, safety and emission of its own section in the presense of a recurrent downstream bottleneck.

\textbf{State $s_t \in S$}: State is a measure of the real time traffic environment, or the
evolution of traffic flow. Due to the complexity of the dynamics of traffic flow, it is quite difficult to obtain a state representation that describes precisely how traffic may change from one state to another. The state variables can be any traffic parameters related to the controlled section that is reported by any sensors such as loop detectors and probe vehicles. As it is suggested in \citep{li2017reinforcement}. The traffic state at the immediate downstream of the merge area, the upstream mainline section, and the on-ramp should be considered for VSL controller. In this paper, the state of the VSL controller agent is defined as $s_t \in R^{m_l+u_l+o_l}$, where $m_l$, $u_l$ and $o_l$ are the numbers of lane of merge area, upstream mainline and on-ramp corrspondingly, the occupancy rate reported by loop detectors are used as state variables. 

\textbf{Action $a_t \in A$}: An action instracts the speed limit of all lanes at time $t$. Therefore $a_t \in R^{c_l}$, where $c_l$ is the number of lane at the controlled section. Considering the real world implementation and the driver compliance issue, the elements of $a_t$ are set as discrete values $a^t_i \in [0,1,2,\cdots, M]$. And the speed limits $V_t \in R^{c_l}$ is equal to $V_0 + I a_t $, where $V_0 \in R^{c_l}$ is the minimum value of the speed limit, $I$ is the integer multiples, the maximum value of speed limits is $V_0 + I M $. For a section with multiple lanes, the dimension of the action space would become very high, hence increase the difficulty of learning. The learning strategy of such action space will be given in the next section.

\textbf{State transition probability $p(s_{t+1}\mid s_t,a_t)$}: The training of agent are conducted on a simulation platform SUMO\footnote{http://sumo.dlr.de}. SUMO provides flexible APIs for network design, traffic sensors and traffic control solutions. The transitions from $p(s_{t+1}\mid s_t,a_t)$ is implicitly defined by SUMO, and the cars in the simulation.

\textbf{Reward $r_t \in \mathcal{R}$}: The agent attempts to maximize its own expected discounted return $E[\sum^{\infty}_{k=0} \gamma^k r_{t+k}]$. Defining a reward function $r_t$ for the VSL control problem is not obvious. The optimized objective of the VSL control can be total travel time, low crash probility, vehicular emission, etc. In order to improve the efficiency of the freeway section, the reward function can not be straightforward defined as average travel time because the travel time of the vehicles cannot be computed until they have completed their routes, which leads to the problem of extremely delayed rewards. Fortunatelly, it is known that there is a direct relationship between the total travel time (TTS) and the inflow and outflow of a traffic network \citep{papageorgiou2002freeway}. The relationship are given as following:
  \begin{equation}
  TTS = KTn_0 + K(K-1)\frac{T^2}{2}(F^{in} - F^{out}),
  \end{equation}
where $K$ is the total time steps. $T$ is the time interval. $F^{in}$ is the total inflow of the transportation network, $F^{out}$ is the total outflow of the network. Obviously, $F^{in} = \sum^\infty_{t=0} f_t^{in}$ and $F^{out} = \sum^\infty_{t=0} f_t^{out}$. Therefore, the reward $r^0_t$ associated with total travel time can be defined as $r^1_t = f_t^{out} - f_t^{in}$. The total outflow $f_t^{out}$ and inflow $f_t^{in}$ at time point $t$ can be easily collected from the loop detectors located on upstream mainline/on-ramp and downstream mainline/off-ramp. \citet{li2017reinforcement} suggested another metric related to the efficiency of freeway. They use the traffic condition of the bottleneck as reward function. The second reward function is considered as the average velocity reported by detectors at bottleneck $r^2_t = avg(vel_t^{bottleneck})$. 

Another objective of VSL control is to reduce crash probility. With SUMO APIs, we can obtain the accleration of the vehicles. The reward $r^3_t$, which is related to the safety of DVSL contolled section, is defined as $r^2_t = -\theta_t$  where $\theta_t$ is the number of emergency braking vehicle (deceleration is above $4.5 m/s^2$) in the last step. Meanwhile, we are also interested in measuring the emission reduction of implementing the DVSL control. SUMO provides flexible APIs to calculate the CO, HC, NOx and PMx emissions. We can use the emission standards to obtain an emission reward $r^4_t$, 
\begin{equation}
r^3_t = -(\frac{e^{CO}_t}{1,5} + \frac{e^{HC}_t}{0.13} + \frac{e^{NOx}_t}{0.04} + \frac{e^{PMx}_t}{0.01})
\end{equation}
where $e^{CO}_t$, $e^{HC}_t$, $e^{NOx}_t$ and $e^{PMx}_t$ are the total CO, HC, NOx and PMx emissions of the freeway section. $1.5$, $0.13$, $0.04$ and $0.01$ are Euro \uppercase\expandafter{\romannumeral6}\footnote{https://eur-lex.europa.eu/legal-content/EN/TXT/?uri=celex$\%$3A32012R0459} standards for CO, HC, NOx and PMx emissions. The usages of four reward functions $r^1_t$, $r^2_t$, $r^3_t$ and $r^4_t$ will be studied in the experiments of section \ref{Ex}.

\section{Deep reinforcement learning for Differential variable speed limit}
\label{DRLDVSL}
\begin{figure}[h!] 
\begin{center}
\includegraphics[width=0.75\textwidth]{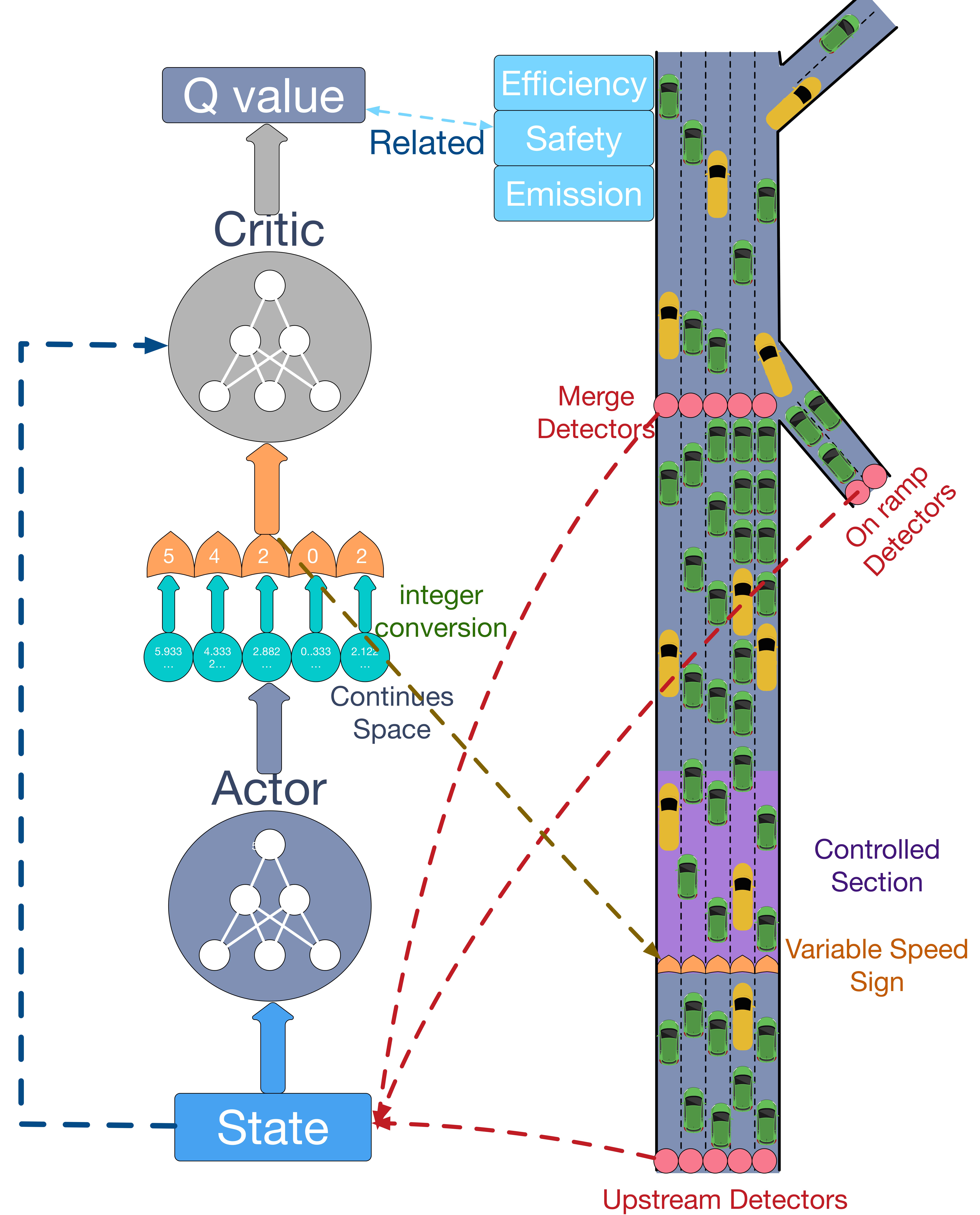}
\caption{The actor-critic architecture for DVSL.}
\label{archi}
\end{center}
\end{figure}

In this section, we present the actor-critic architecture for DVSL. The basic framework of the architecture is given in Fig. \ref{archi}. This architecture avoids the
heavy cost of evaluating all sets of different speed limits among lanes. This policy builds upon the actor-critic framework. The actor is used to generate action for VSL control, and the critic is utilized to evaluate the actor’s policy. The estimated Q value of critic is related to the efficiency, safety and emission of the transportation network. Multi-layer neural networks are used as function approximators for both the actor and critic functions.

\subsection{Action Generation}
The architecture reasons over actions within a continuous space $R^{c_l}$. and then simply using the integer conversion to generate the discrete action $a_t$. The speed limits can be obtained by $V_0 + I a_t $. The input state $s_t \in R^{m_l+u_l+o_l}$ of the actor is the occupancy rates collected from the upstream, merge area and on-ramp detectors. The actor can be defined as following:
\begin{equation}
 \begin{split}
f_{\theta^\pi}: R^{m_l+u_l+o_l} \rightarrow R^{c_l},\\
f_{\theta^\pi}(s_t) = \widehat{a_t}.
 \end{split}
\end{equation}
$f_{\theta^\pi}$ is a function parameterized by $\theta^\pi$, mapping from state space $R^{m_l+u_l+o_l}$ to action space $R^{c_l}$. As stated before, the continues speed limits is not feasible to post in the variable speed sign. As a result, the speed limit calculated from $\widehat{a_t}$ will likely not be a valid one. Suppose that there are $M$ kinds of speed limits for each lane, we need to be able to map from $\widehat{a_t}$ to an element in the valid action set $A$. The proposed mapping strategy is given as following: 
\begin{equation}
 \begin{split}
g: R^{c_l} \rightarrow A,\\
g(\widehat{a_t}) = int(clip(\widehat{a_t}, 0, M)).
 \end{split}
 \label{v_gene}
\end{equation}
$g$ is a mapping from a continuous space to a discrete set. It first clips the values of $\widehat{a_t}$ into $(0, M)$, then the discrete action $a_t$ can be easily obtained by the integer parts of clipped $\widehat{a_t}$. It guarantees the values of $a_t \in (0,1,\cdots,M-1)$. The different speed limits among lanes can be then calculated by $V_t = V_0 + a_tI$. The work of \citep{dulac2015deep} uses a k-nearest neighbor to map from continues action into discrete action, which has to be performed in logarithmic time. It is obvious that our approach is more efficient than the k nearest neighbor approach for DVSL control problem.  

The critic function is used to evaluate the action representation, and estimate the value function  $Q_{\theta^Q}$ of choice of actor:
\begin{equation}
\pi_\theta(s) = \argmax_{a \in g(f_{\theta^\pi}(s))} Q_{\theta^Q}(s,a),
\end{equation}
where $Q_{\theta^Q}$ is the estimated value function parameterized by $\theta^Q$, $\theta^Q$ represents the parameter of the critic, $\theta^\pi$ represents the parameter of the actor. The goal of actor for VSL control is to maximize the value function $Q$.
\begin{equation}
Q(s_t, a_t) = E[\sum^{\infty}_{k=0} \gamma^k r_{t+k}(s_t, a_t)] = r_t + \gamma^k Q(s_{t+1},a_t),
\end{equation}
where $Q$ is the actual value function, $r_t$ is the reward function, which is depending on the reward selection for VSL control system. The goal of critic is to produce a perfect approximation of value function. For a complex MDP, the true value function can not be obtained until a large number of policy have been tried. But it can be learn by bootstrapping from the current estimate of the value function. The parameters $\theta^Q$ of the critic can be updated by the temporal difference (TD) error signal:
\begin{equation}
\acute{\theta^Q} = argmin_{\theta^Q} Q_{\theta^Q}(s_t,a_t) - (r(s_t, a_t) + Q_{\theta^Q}(s_{t+1},\pi_\theta(s_{t+1}))),
\end{equation}  
where $\delta = Q_{\theta^Q}(s_t,a_t) - (r(s_t, a_t) + Q_{\theta^Q}(s_{t+1},a_t))$ is the TD error signal. 

\subsection{Training with Deterministic Policy Gradient}
The training algorithm’s goal is to find a parameterized policy $\pi_\theta$ which maximizes its expected reward return over the DVSL controlled time period. $\pi_\theta$ is characterized by the actor $g(f_{\theta^\pi})$. We perform the training using Deep Deterministic Policy Gradient (DDPG) \citep{lillicrap2015continuous}. The core of DDPG is to use a stochastic behavior policy for exporation but estimate a deterministic target policy, which means that the final estimated $\pi_\theta(s_t)$ will produce a deterministic action $a_t$ rather a stochastic one. The DDPG optimizes the parameters of actor-critic in a bilevel optimization manner, the loss function $L(Q,\theta^Q)$ for critic is:
\begin{equation}
 \begin{split}
y_i = r(s_i, a_i) + Q_{\grave{\theta^Q}}(s_{i+1}, \pi_{\grave{\theta^\pi}}(s_{i+1})),\\
L(Q,\theta^Q) = \frac{1}{N} \sum_i (y_i - Q_{\theta^Q}(s_i,a_i))^2.
 \end{split}
 \label{criticloss}
\end{equation}
Here $N$ is the number of samples, with the $i$ index referring to the $i$th sample. $y_i$ is the $i$th label computed from the sum of the immediate reward and the outputs of the target actor and critic networks, having weights $\grave{\theta^\pi}$ and $\grave{\theta^Q}$ respectively. Then the critic loss $L(Q,\theta^Q)$ can be computed. The weights $\theta^Q$ of the critic network can be updated with the gradients obtained from the loss function in Eq \ref{criticloss}. The goal of the actor is to optimize the loss function:
\begin{equation}
L(\pi, \theta^\pi) = - \frac{1}{N} \sum_i Q(s_i, \pi_{\theta^\pi}(s_i)).
\end{equation}
In order to update the weights $\theta^\pi$, we can use the gradient:
\begin{equation}
\bigtriangledown_{\theta^\pi} = \frac{1}{N} \sum_i \bigtriangledown_a Q_{\theta^Q}(s_i, a) \mid_{a = \pi_{\theta^\pi}(s_{i})} \bigtriangledown_{\theta^\pi} \pi_{\theta^\pi}(s_{i}).
\label{actorloss}
\end{equation}
The detail of the deterministic policy gradient in Eq \ref{actorloss} can be found in \citep{silver2014deterministic}, it is proved that the deterministic policy gradient in Eq \ref{actorloss} is equivalent to the stochastic policy gradient. If we take $g$ as parts of the actor, the architecture of the actor will not be fully differentiable. However, $g$ can be considered as a function of the VSL signs. The VSL signs can take the action generated from $f_{\theta^\pi}$, and use $g$ to produce feasible speed limits. 

Traditional RL agents incrementally sample the experience including state $s_i$, $s_{i+1}$, action $a_i$ and reward $r_i$ to update their parameters and discard those experience immediately. This approach will cause strong temporal correlation between samples and rapid forgetting of possibly useful experiences. Experience replay \citep{lin1992self,mnih2015human} addresses both of these problems by storing experience into a replay memory. The experience are constantly sampled from the replay memory to update the agents, which stabilized the training of neural networks for DRL. In this paper, we apply priority experience replay proposed in \citep{schaul2015prioritized} to sample experience to update the actor weights $\theta^\pi$ and critic weights $\theta^Q$, in which the probability $p_i$ of transition $(s_i, a_i, r_i, s_{i+1})$ being sampled from replay memory is:
\begin{equation}
p_i = \frac{1}{rank(i)}, 
\end{equation}
$rank(i)$ is the rank of transition i when the replay memory is sorted according to new or old degree. The central idea is that new experience is more valuable.

A core challenge in RL is how to balance exploration--actively seeking out actions that might yield high rewards and lead to long-term gains; and exploitation--maximizing short-term rewards using the agent’s current knowledge. Without adequate exploration, the agent might fail to discover effective DVSL control strategies. One advantage of DDPG, as an off-policy RL framework, is that its exploration can be independent from the learning algorithm. The exploration is done by adding noise $x$ sampled from a noise process to $f_{\theta^\pi}(s_t)$. In the experiments, the noise $x$ is modeled as a laplace process $L(x \mid b) \sim \frac{1}{2b_t} exp(-\frac{x}{b_t})$. The parameter $b_t$ is decay with respect to the learning time. The algorithm for framework of DDPG for DVSL control is summarized in algorihtm.1.

\begin{algorithm}[htb] 
\caption{ DVSL control agent training with DDPG.} 
\small
\label{alg:Framwork} 
\begin{algorithmic}[1]
\STATE {Set a reward function that is chosen from $r^1, r^2, r^3, r^4$ and the the integer multiples $I$ for DVSL.}
\STATE {Randomly initialize critic network $Q_{\theta^Q}$ and actor network $\pi_{\theta^\pi}$ with parameters $\theta^Q$ and $\theta^\pi$;}
\STATE {Initialize target weights: $\grave{\theta^\pi} \to \theta^\pi$, $\grave{\theta^Q} \to \theta^Q$.}
\STATE {Initialize replay memory.}
\FOR{$episode=1$ to $m$}
\STATE {Start the traffic simulation with SUMO.}
\STATE {Initialize a random process $L(x \mid b) \sim \frac{1}{2b_t} exp(-\frac{x}{b_t}) $ for action exploration.}
\STATE {Recieve initial observe state $s_1$ from the loop detectors in SUMO.}
\FOR{$t=1$ to time length of the traffic simulation $T$}
\STATE {Select action $a_t = g(f_{\grave{\theta^\pi}}(s_t)+x_t)$ according to the current policy and exploration method.}
\STATE {Decay the noise parameter $b_t$.}
\STATE {Execute DVSL with speeds $V_0 + I*a_t$ and observe reward $r_t$, new state $s_{t+1}$ from the SUMO simulation.}
\STATE {Store $(\{t, s_t , a_t , r_t, s_{t+1}\})$ in the replay memory according to the rank of new and old degree.}
\STATE {Sample a random minibatch of $k$ transitions $(\{i, s_i , a_i , r_i, s_{i+1}\})$ from the replay memory using probility $\frac{1}{rank(i)}$;}
\STATE {Update the critic by minimizing the loss $L(Q,\theta^Q)$ in \ref{criticloss}.}
\STATE {Update the actor $f_{\theta^\pi}$ by sampled gradient given in \ref{actorloss}.}
\STATE {Update the target network: 
\begin{equation*}
\begin{split}
\grave{\theta^\pi} \gets \tau \theta^\pi + (1 - \tau) \grave{\theta^\pi},\\
\grave{\theta^Q} \gets \tau \theta^Q + (1 - \tau) \grave{\theta^Q}
\end{split}
\end{equation*}}
\ENDFOR
\ENDFOR
\end{algorithmic}
\end{algorithm} 

\section{Experiment}
\label{Ex}
\subsection{The traffic network in SUMO as the environment of the agents}
\begin{figure}[h!] 
\begin{center}
\includegraphics[width=0.55\textwidth]{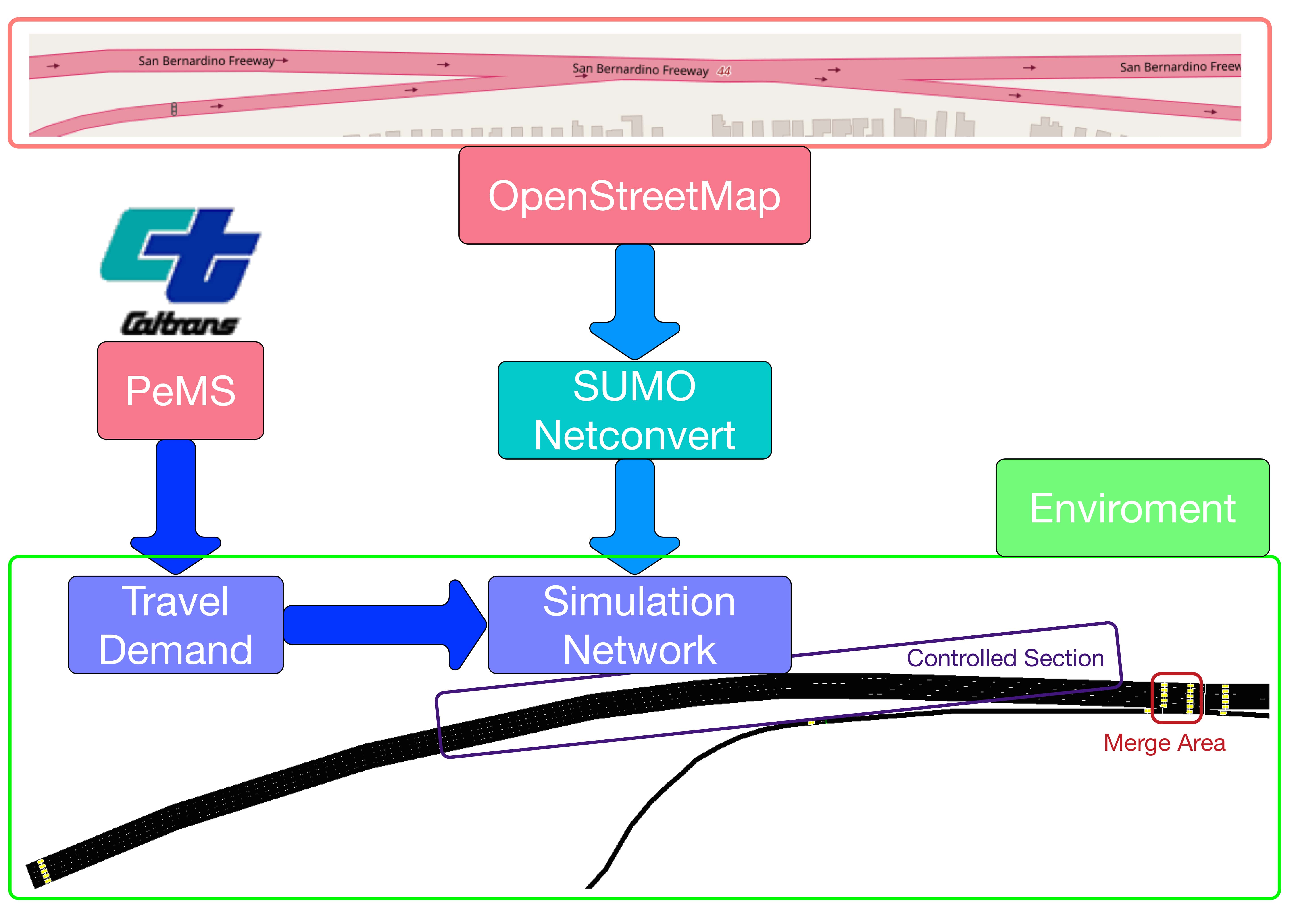}
\caption{The simulation enviroment in SUMO.}
\label{simul}
\end{center}
\end{figure}

Open source software SUMO is selected for the experiments. The software is highly flexible, well documented and supports set the speed limits for each lane using its API--the Traffic Control Interface (TraCI) package. A 5 lane section with on/off ramps of San bernadino freeway located in State of California is selected. A map of the section is first exported from OpenStreetMap.org. Next, the map is used to generate the traffic network for simulation. It should be noted the traffic network can be any freeway section with on-ramp and/or off-ramps. The goal of this paper is to evaluate the usage of deep reinforcement learning on DVSL. The reason why we choose this particular freeway section is its simple structure. Moreover, we can use data from an open datasets--California PeMS\footnote{http://pems.dot.ca.gov} to generate demand data for simulation. The simulation enviroment is summarized in Fig \ref{simul}. The length of the merge area with recurrent bottlenecks is $26.87m$, the length of controlled section is $780.35m$. The original speed limits for mainlane and on/off ramps are 65 mph and 50 mph respectively.

\begin{figure}
  \centering
  \subfigure[The hourly demand for simulation 1]{
    \label{fig:subfig:a} 
    \includegraphics[width=0.41\textwidth]{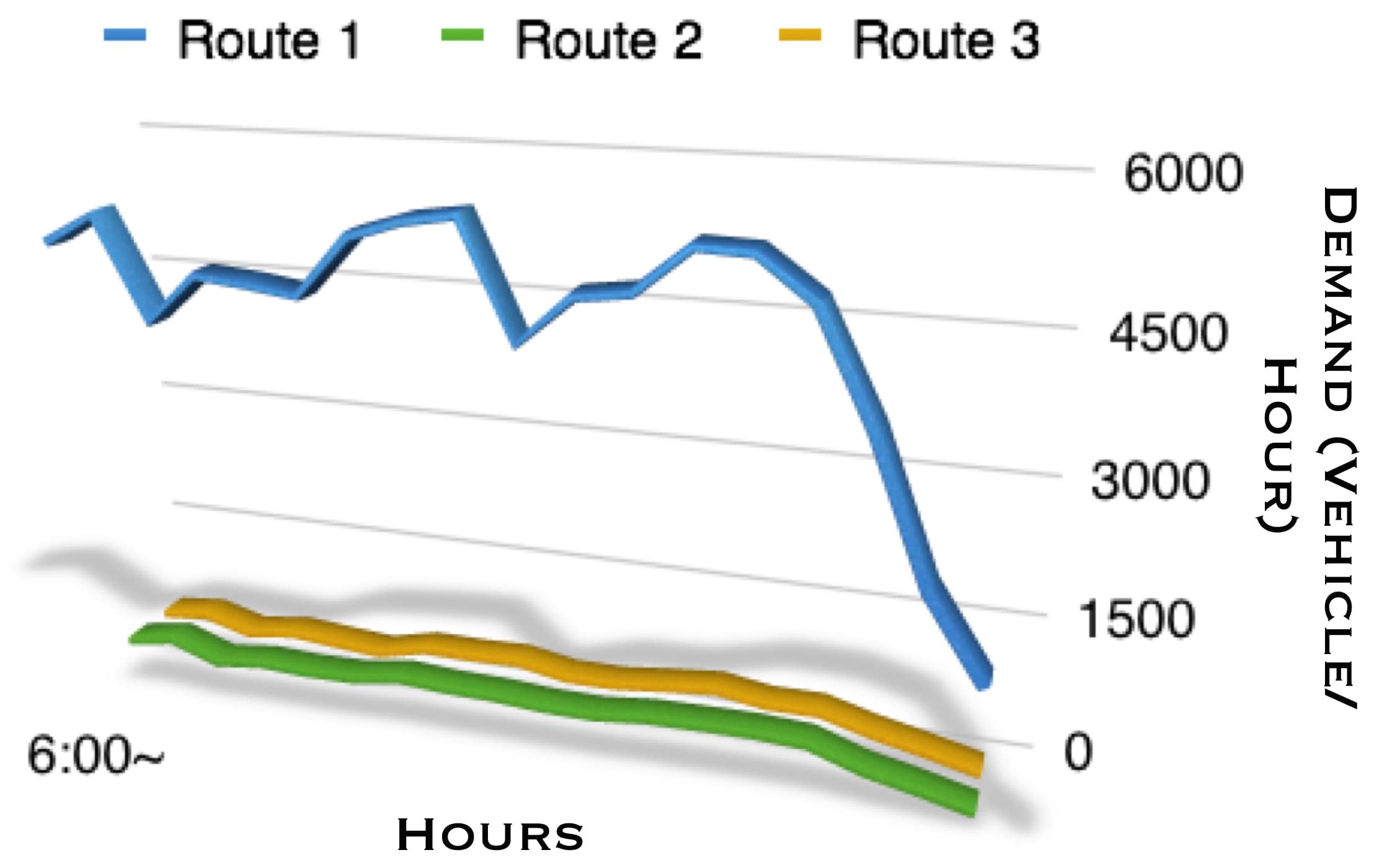}}
  \hspace{0.0in}
  \subfigure[The average 5 min speed collected from detectors in merge area without VSL control for simulation 1]{
    \label{fig:subfig:b} 
    \includegraphics[width=0.41\textwidth]{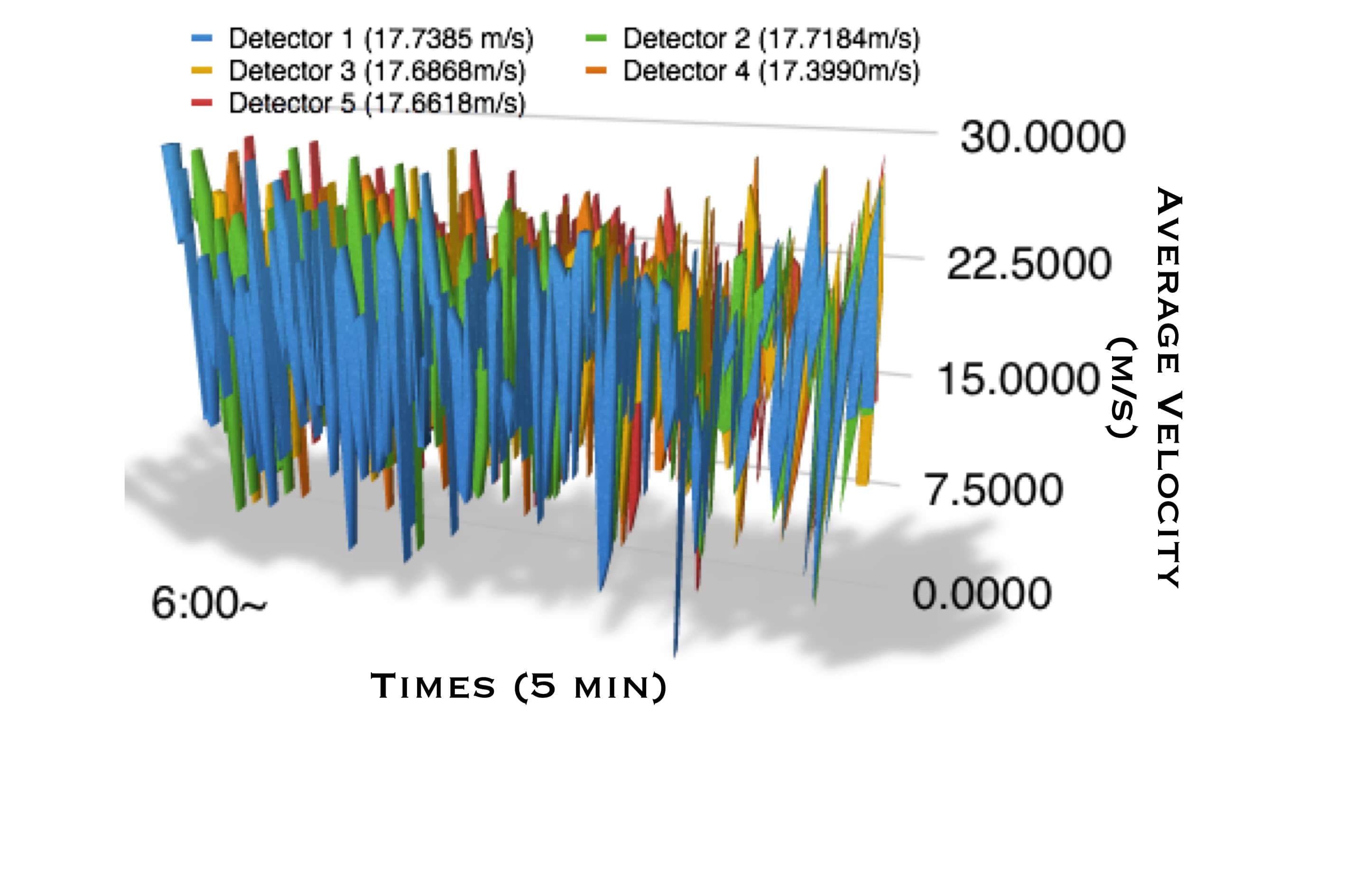}}
  \label{fig:subfig} 
  \hspace{0.0in}
    \subfigure[The hourly demand for simulation 2]{
    \label{fig:subfig:c} 
    \includegraphics[width=0.41\textwidth]{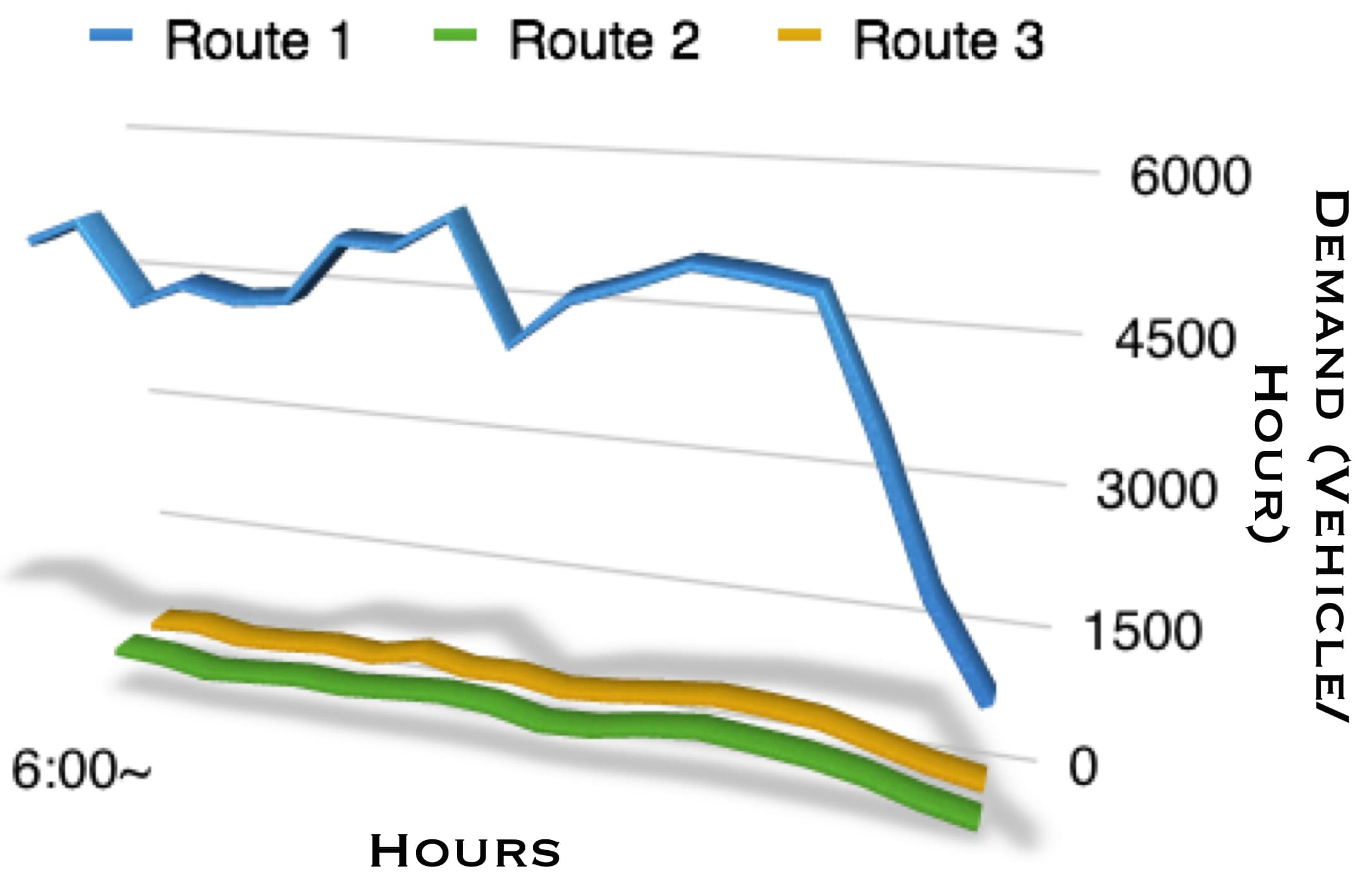}}
  \hspace{0.0in}
  \subfigure[The average 5 min speed collected from detectors in merge area without VSL control for simulation 2]{
    \label{fig:subfig:d} 
    \includegraphics[width=0.41\textwidth]{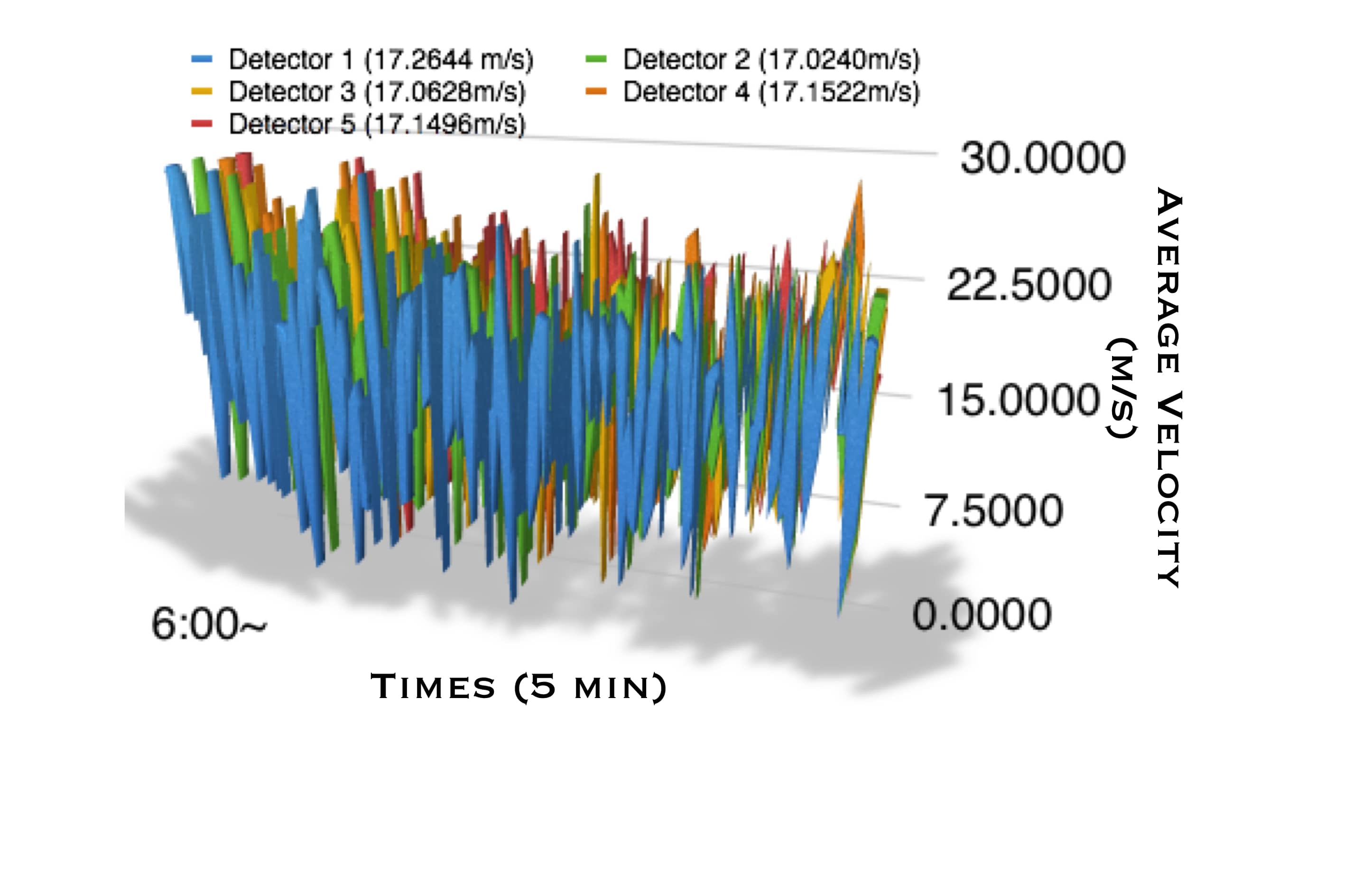}}
  \label{fig:subfig} 
  \caption{The hourly demands and average 5-min velocities collected from detectors in merge area.}
  \label{demand_speed} 
\end{figure}

Next, we need to generate travel demand data for the simulation. For the networks in Fig \ref{simul}, we need to consider only 3 route choices: 1) From mainlane to mainlane, 2) From mainlane to off-ramp and 3) From on-ramp to mainlane. The station with ID 717241 from PeMS is near the freeway section. Based on some observations from traffic flow recorded by detectors of station 717241, simulation demand is generated to approximate the routes into the system. The demand is purposefully set to cause recurrent bottlenecks. Each round simulation lasts for 18 hours from 6:00 to 24:00. The numbers of vehicles with 3 routes in each hour are modeled as Poisson process. Two vehicle types: passenger car with length 3.5m and truck/bus with length 8m, are considered, the percentage of passenger vehicle is 85\%, and the percentage of truck/bus is 15\%. A new shedule of demand is randomly generated in each round simulation according to the ramdom distributions. Fig \ref{demand_speed} gives two examples of random demands and average speeds collected from loop detectors in the merge area without VSL control. It can be found that the merge area is easy to suffer traffic breakdown with the given traffic demand.

\subsection{Action space and parameters for the agent}
The proposed DVSL control strategy adjusted speed limit from 50 to 75 mph with an increment of 5 mph. The action set for each lane is given by [50,55,60,65,70,75] mph ([22.45, 24.695, 26.94, 29.185, 31.43, 33.679] m/s). There are 6 options for the agent, therefore the number $M$ of $g$ in Eq \ref{v_gene} is set as 6. The state dimension for the agent is 11. There are 5 lanes in its upstream section, 5 lanes in its downstream merge area and 1 lane in the on-ramp. The actor and critic of the agent are composed of neural networks. The actor $f_{\theta^\pi}$ is expressed by:
\begin{equation}
\begin{split}
h^\pi_t = relu(W^\pi_1s_t + b^\pi_1),\\
a_t = M \times sigmoid(W^\pi_2h^\pi_t + b^\pi_2)
\end{split}
\end{equation} 
where $relu$ and $sigmoid$ are nonlinear activations. $W^\pi_1 \in R^{120 \times 11}$, $b^\pi_1 \in R^{120}$, $W^\pi_2 \in R^{5 \times 120}$ and $b^\pi_2 \in R^{5}$ are the parameters $\theta^\pi$ for the actor. The critic $Q_{\theta^Q}$ is expressed by:
\begin{equation}
\begin{split}
h^Q_t = relu(W^Q_ss_t + W^Q_aa_t + b^Q_1),\\
Q_t = (W^Q_2h^Q_t + b^Q_2)
\end{split}
\end{equation}
where $W^Q_s \in R^{120 \times 11}$, $W^Q_s \in R^{120 \times 5}$, $b^Q_1 \in R^{120}$, $W^\pi_2 \in R^{1 \times 120}$ and $b^Q_2 \in R$ are the parameters $\theta^Q$ for the critic. During training, the noise parameter $b_t$ for exploration is set to $2.5$, and decays 99.9\% in each step. We train 4 actor-critic models using rewards $r^1$, $r^2$, $r^3$ and $r^4$ given in Sec \ref{MDPF} and conduct comparison between them. The action are dynamically change in 1 minute, which means that the speed limits of the controlled area change every minute.

\subsection{Evaluation}
After the training of the models are finished, the simulations are run 50 episodes to report the experimental results. To make sure the comparison is fair, all models are evaluated on simulations with the same demand. All the RL based models are trained with 150 episodes before evaluation, the parameters of all models are fixed after training. The average accumulated reward values and average travel time (ATT) are calculated for quantitative evaluation. The average CO, NOx, HC and PMx emissions are also provided. The 4 DDPG based DVSL controllers are benchmarked against the baseline scenario in which no control occurs at all, and baselines with Q learning and DQN based VSL controllers:
\begin{itemize}
\item[•] \textbf{NoVSL, baseline}: The baseline without any VSL. The vehicles are running with speed limits 65 mph in the mainline.
\item[•] \textbf{Q learning, same speed limit for each lane}: A similar model proposed by \citet{li2017reinforcement}. The Q table size of the agent is $125 \times 6$. The average occupancy in upstream section, merge area and on-ramp over lanes are used to determine the states for the agent. $r^1$ is used as the reward to train the agent.
\item[•] \textbf{Deep Q networks (DQN), same speed limit for each lane}: A deep version of Q learning. The state of the agent is as the same as the DDPG models. The neural network structure of the DQN model is:
\begin{equation}
\begin{split}
h^{QN}_t = relu(W^{QN}_1 s_t + b^{QN}_1), \\
Q(s_t, a_t) = (W^{QN}_2 h^{QN}_t + b^{QN}_2).
\end{split}
\end{equation} 
where $W^{QN} \in R^{120 \times 11}$, $b^{QN}_1 \in R^{120}$, $W^{QN}_2 \in R^{M \times 120}$ and $b^{QN}_2 \in R^M$. This parameters guarantees that the DQN model have similar learning capability with the DDPG models. As the same as Q learning based models, $r^1$ is used as the reward to train the agent. More detail for learning DQN models can be found in \citep{mnih2015human}.
\end{itemize}

\subsection{Results}
Table \ref{table1} compares the average performance of different RL over the 50 episodes of evaluated simulation by different indexes. At first glance, it is evident that DDPG based DVSL control strategies outperform Q learning based VSL approaches and NoVSL condition in terms of almost all the indexes. DDPG-$r^1$ is the best controller in terms of the efficiency of the transportation network, it reduces nearly 40 seconds average ATT per episode. From the viewpoint of safety, DDPG-$r^3$ causes the lowest number of emergency braking vehicles, whereas other models do not improve the $r^3$ index compared with NoVSL control. From the environmental standpoint, DDPG-$r^4$ is the
best controller because it leads to the lowest Co, Hc, Nox and Pmx pollution, as well as the highest $r^4$ value. Moreover, DDPG-$r^3$ is also a valuable controller in terms of emission.

\begin{table}[htbp] \tiny
\centering  
 \caption{Average performance of different models on one episode of simulation. The best controller for each index are shown in boldface.}
\begin{tabular}{c c c c c c c c c c} 
\toprule[2pt]
method & $r^1$($10^3$) & $r^2$($10^3$) & $r^3$ ($10^3$) & $r^4$($10^7$) & Co(Kg) & HC(Kg)  & Nox(Kg) & Pmx(Kg) &ATT(s) \\
\midrule[1pt]
NoVSL      &-1.407   &4.645  & -1.872 &-2.952  &2860  &14.70  &30.85  &1.616  &346.3 \\
Q learning &-1.392   &4.763  & -1.866 &-2.915  &2809  &14.53  &30.91  &1.613  &321.1 \\
DQN        &-1.389   &4.812  & -1.912 &-2.920  &2811  &14.57  &30.93  &1.616  &318.6 \\
DDPG-$r^1$ &\textbf{-1.360}   &4.991  & -1.918 &-2.850  &2735  &14.11  &30.36  &1.592  &\textbf{306.4} \\
DDPG-$r^2$ &-1.362   &\textbf{5.067}  & -1.873 &-2.912  &2805  &14.45  &30.80  &1.615  &327.8 \\
DDPG-$r^3$ &-1.398   &4.337  & \textbf{-1.731} &-2.812  &2684  &13.84  &30.23  &1.579  &319.3 \\
DDPG-$r^4$ &-1.427   &4.887  & -1.852 &\textbf{-2.794}  &\textbf{2673}  &\textbf{13.82}  &\textbf{29.73}  &\textbf{1.546}  &317.9 \\
\bottomrule[2pt]
\end{tabular}
 \label{table1}
\end{table}

The Q learning model and DQN model are trained with the reward signal $r^1$, they achieve worse performance compared with DDPG-$r^1$. The significant performance difference between VSL and DVSL controllers indicates the potential value of DVSL in application. However, it should be noted that the lane changing behavior of the vehicles in SUMO are very rational, there will always be a reality gap between the simulation and reality. The usage of DVSL in real world freeway still requires further studies.  

From Table \ref{table1}, it can be seen that each DDPG model achieves the best accumulated reward that it is trained with. Comparing 4 reward signals, we can find some interesting findings: 1) The DDPG-$r^1$ controller is better than the DDPG-$r^2$ one though both $r^1$ and $r^2$ are related to the efficiency of the freeway. The reason is that the reward signal $r^2$ simply averages speed over all lanes of the merge area, it neglects many important information including upstream/downstream condition, on/off ramp information and speed difference between lanes. While the reward signal $r^1$ is directly related to the total travel time of the freeway. The experimental results suggest that $r^1$ is a more powerful reward signal compared with $r^2$. 2) The reward signal $r^3$ is the best one considering all the indexes. It significantly improves the safety by reducing the number of emergency braking vehicle, and its emission and congestion index are all relatively low. But it should be noted that the reward signal $r^3$ is currently difficult to obtain, it is calculated by the accleration of all vehicles in the controlled section. It is impossible to obtain such reward without a highly developed CAV enviroment. 3) The reward signals related to efficiency, safety and emission are highly coupled. For example, the DDPG controller trained to reduce congestion can reduce the emission and vice versa. If we can find the common structure of the reward signals, a more powerful DVSL controller can be built to improve all the aspects of the transportation network.

\subsection{Discussion}
\begin{figure}[htbp]
  \centering
  \subfigure[The speed limits among different lanes during 7:00--7:59 generated from DDPG-$r^1$]{
    \label{pa} 
    \includegraphics[width=0.31\textwidth]{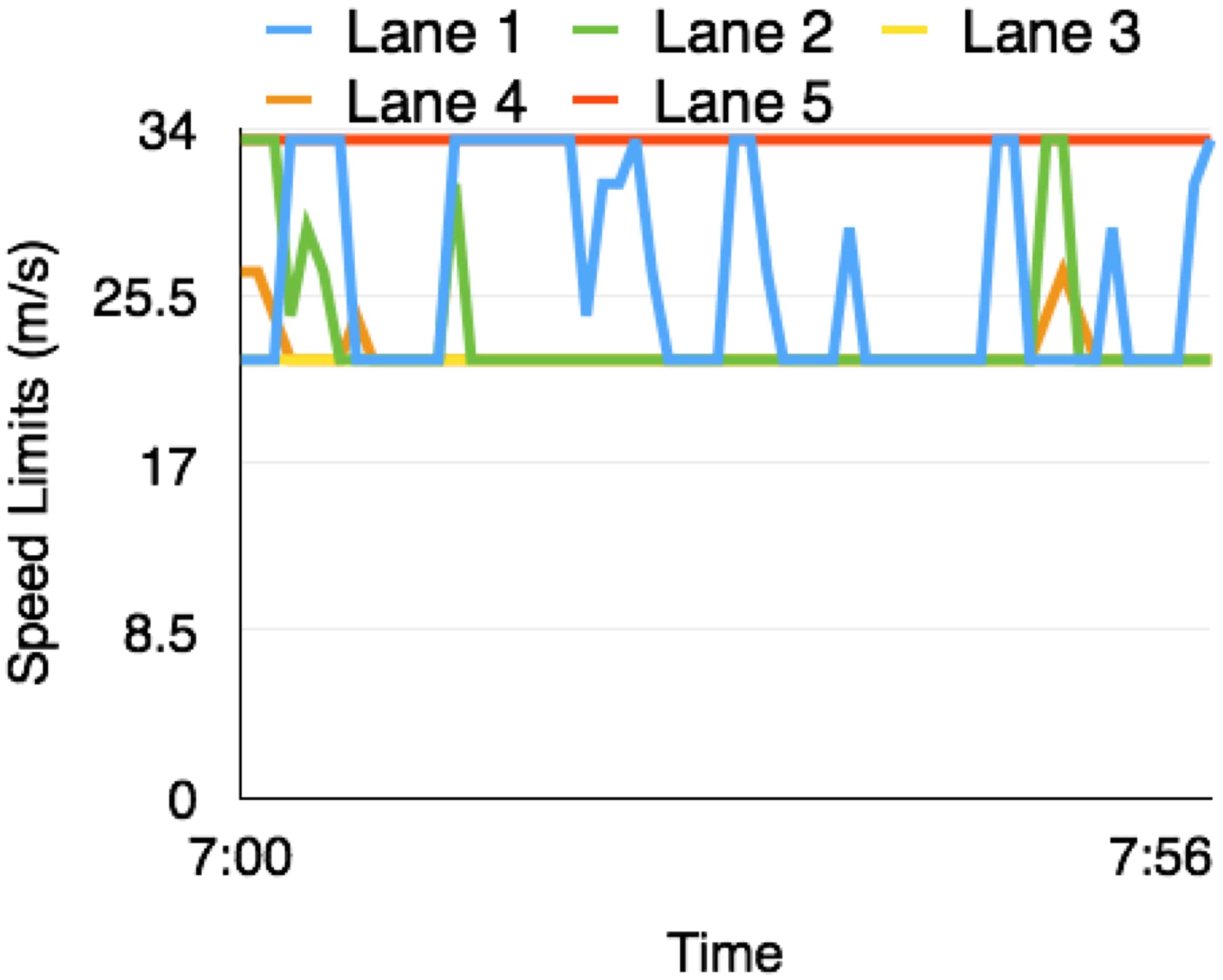}}
  \hspace{0.0in}
  \subfigure[The speed limits among different lanes during 7:00--7:59 generated from DDPG-$r^2$]{
    \label{pb} 
    \includegraphics[width=0.31\textwidth]{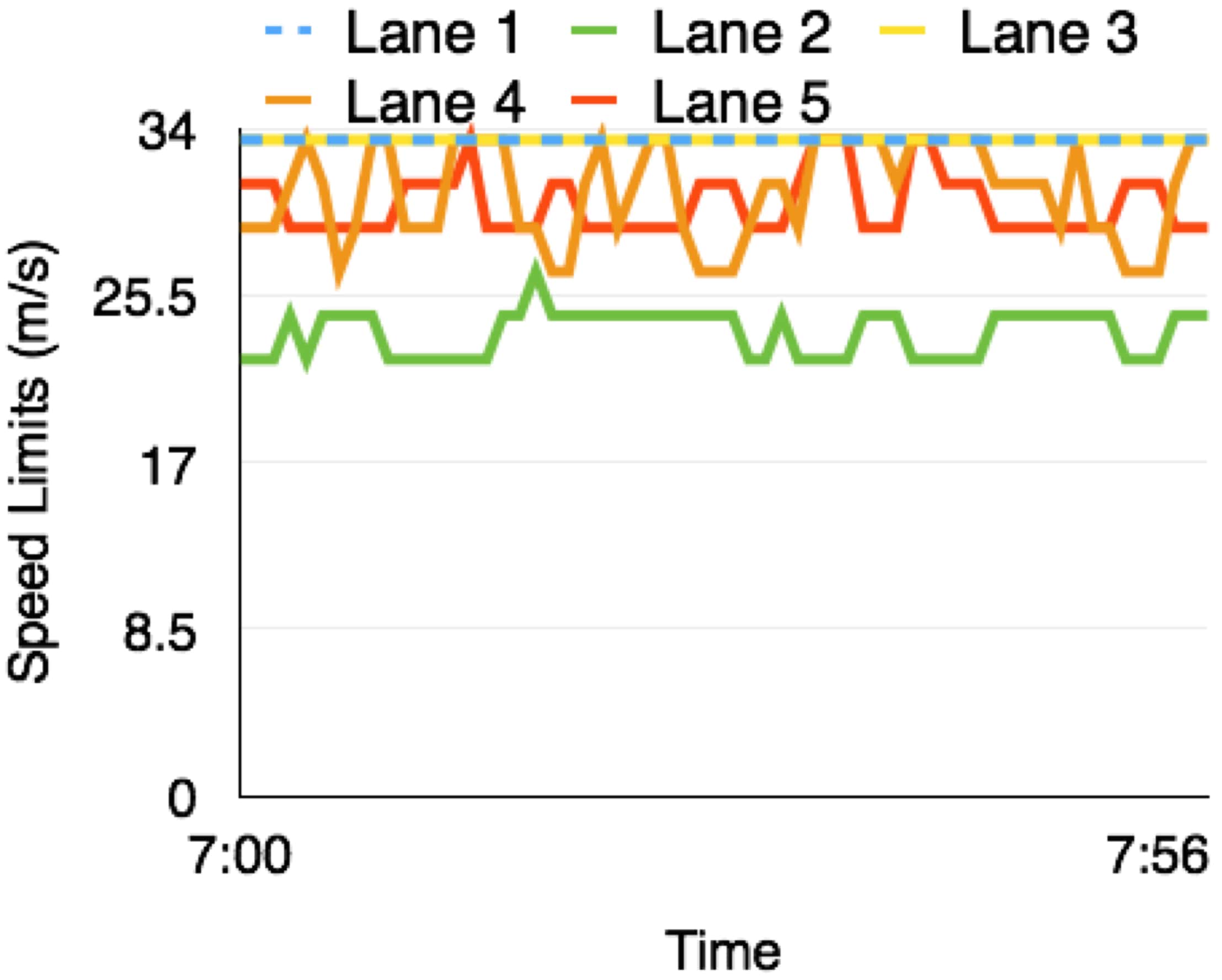}}
  \label{fig:subfig} 
  \hspace{0.0in}
    \subfigure[The speed limits among different lanes during 7:00--7:59 generated from DDPG-$r^3$]{
    \label{pc} 
    \includegraphics[width=0.31\textwidth]{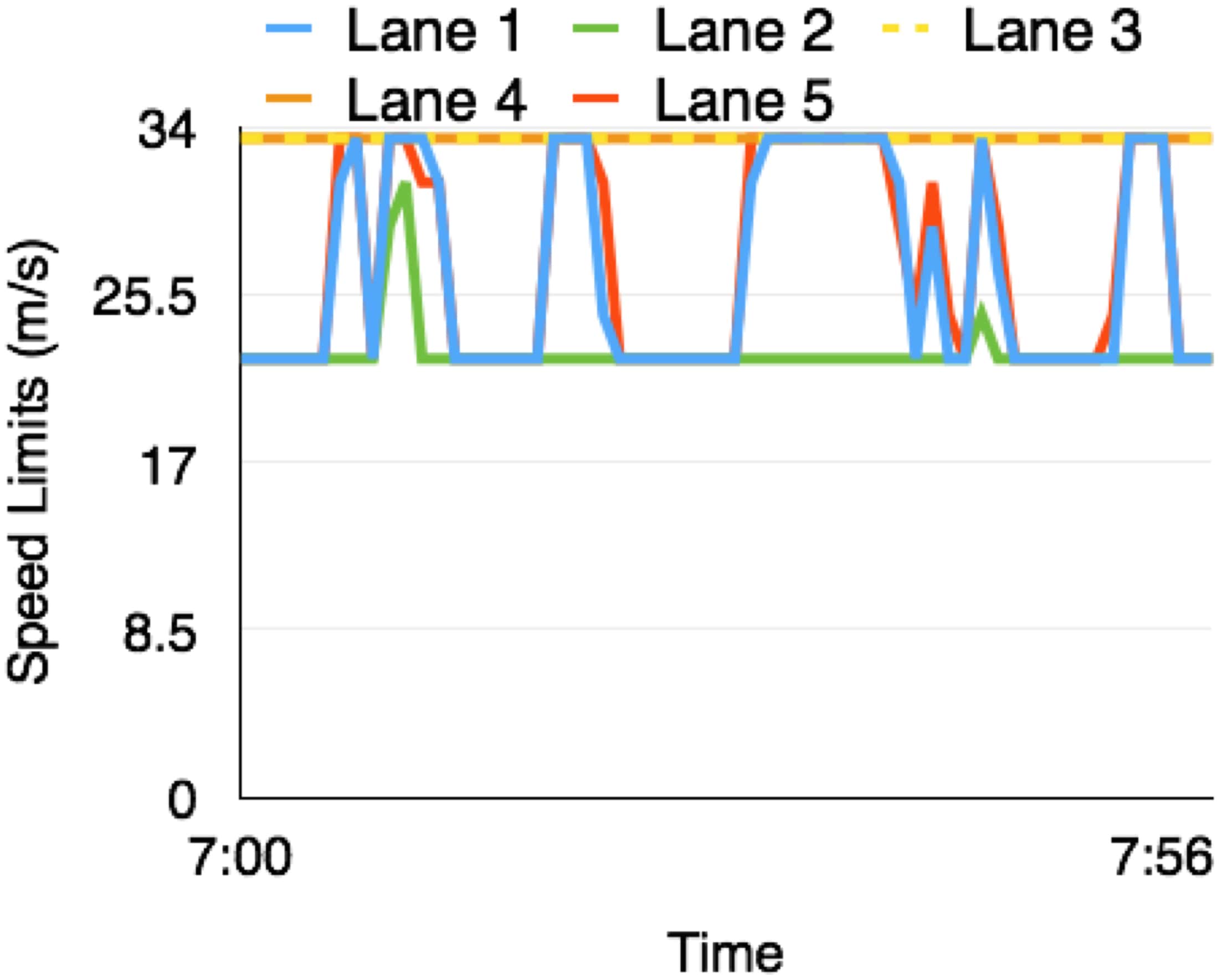}}
  \hspace{0.0in}
  \subfigure[The speed limits among different lanes during 7:00--7:59 generated from DDPG-$r^4$]{
    \label{pd} 
    \includegraphics[width=0.31\textwidth]{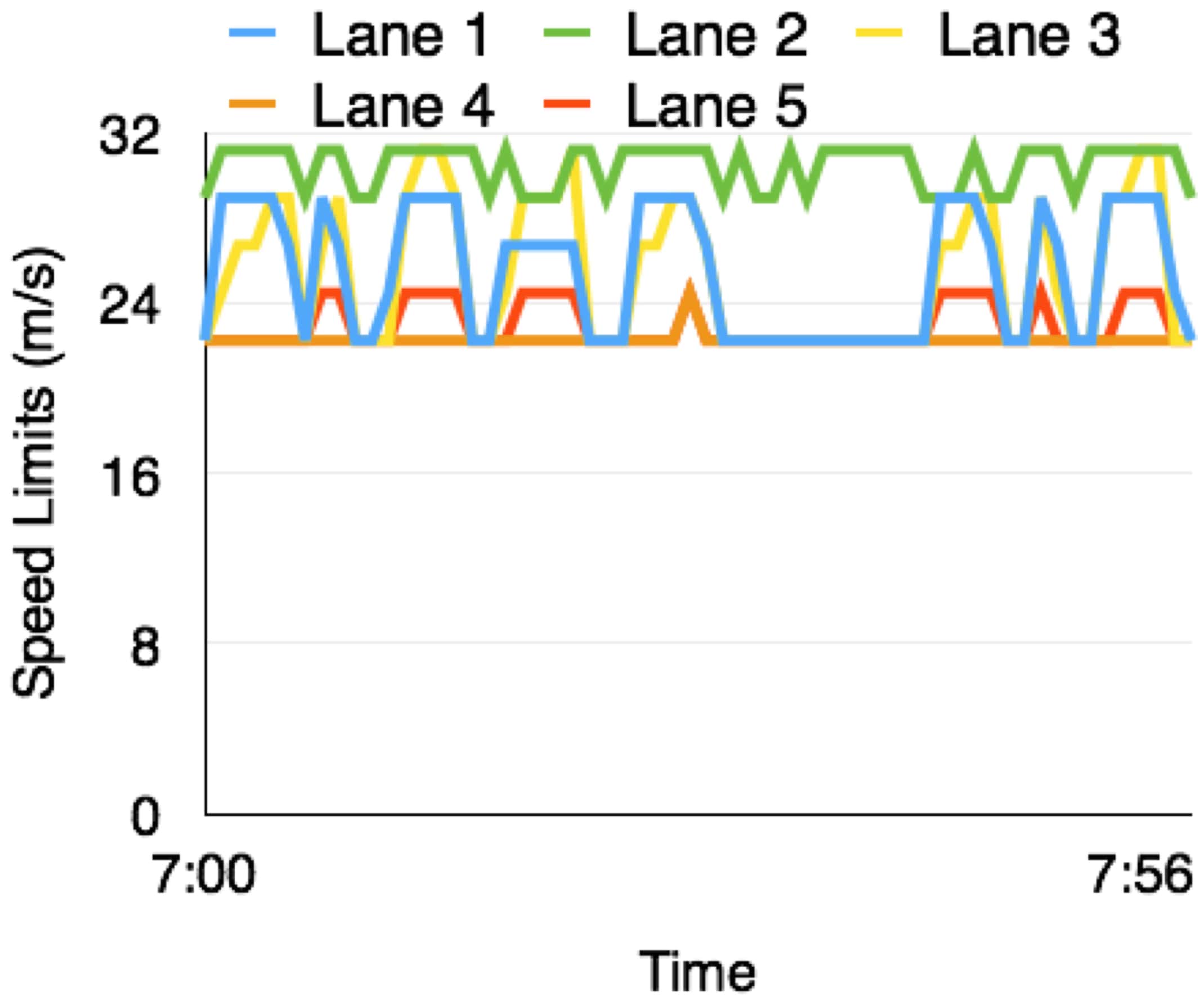}}
     \hspace{0.0in}
    \subfigure[The speed limits during 7:00--7:59 generated from DQN]{
    \label{pe} 
    \includegraphics[width=0.31\textwidth]{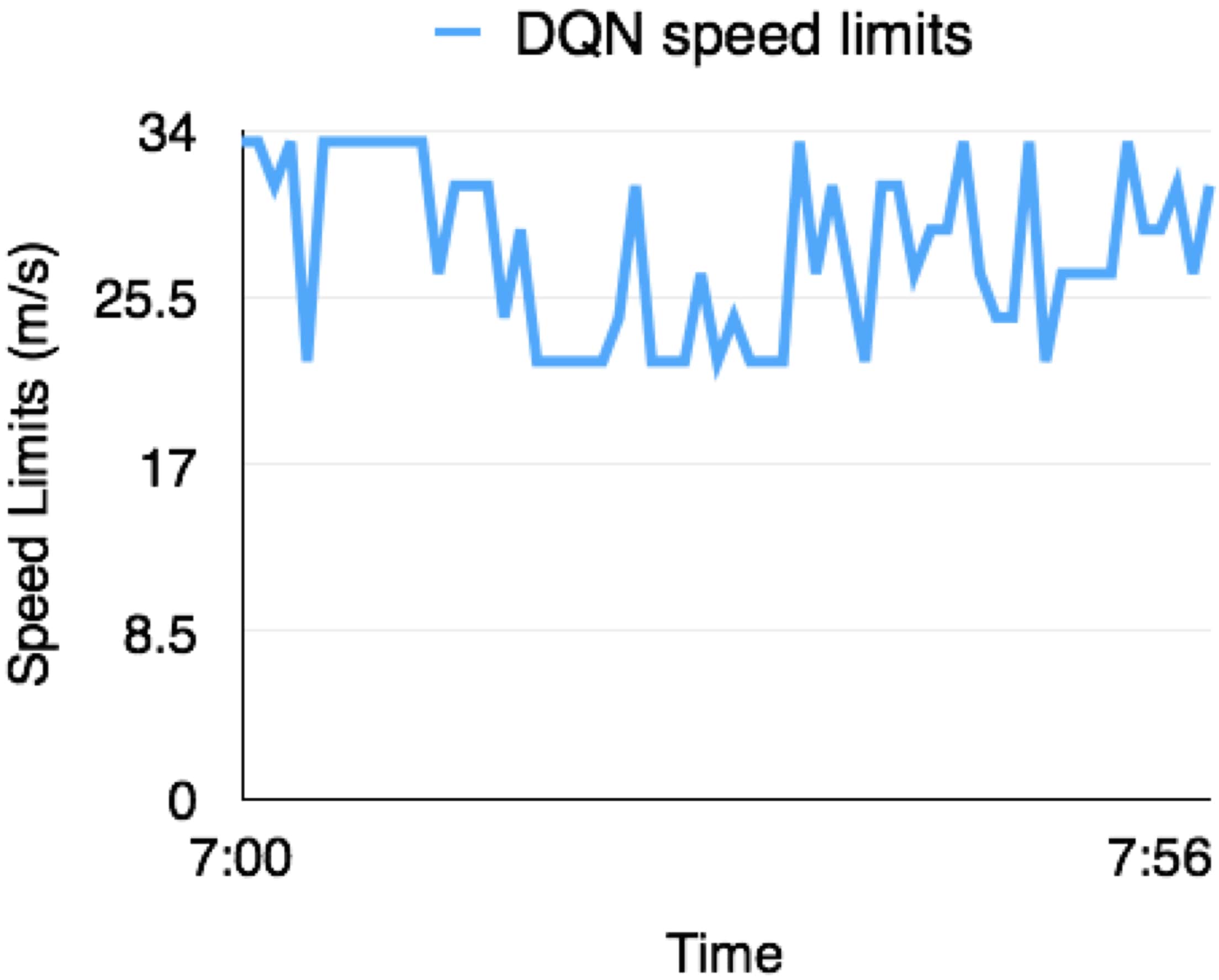}}
  \hspace{0.0in}
  \subfigure[The correlation coefficient between speed limit policies generated from different DRL models]{
    \label{pf} 
    \includegraphics[width=0.31\textwidth]{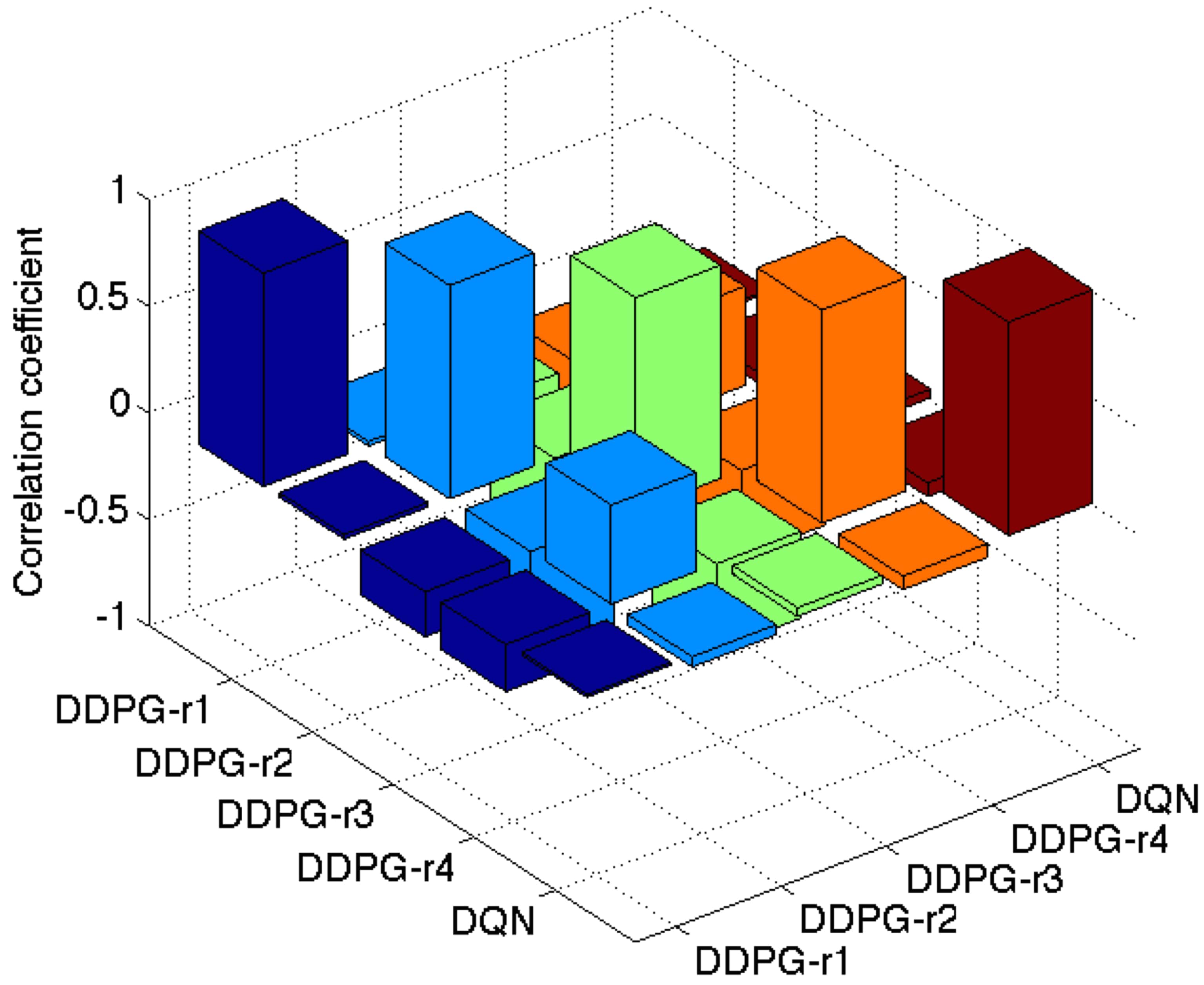}}
  \caption{Speed limit variation generated from different DRL models and correlation coefficients between them}
  \label{policyplot} 
\end{figure}

For traffic control problem, it is very important to study the control policies generated from data-driven models rather than simply showing the overall evaluation indexes. To further understand how the DVSL policies generated from different DRL models are different from each other, we plot a small period speed limit variations from an evaluated simulation run of the proposed 4 DDPG based controllers, as well as the DQN based VSL controller. Fig. \ref{pa}--\ref{pe} show the speed limits generated from different DDPG models, it is obvious the policies of DDPG trained by different rewards are significantly different from each other. Furthermore, Fig. \ref{pf} shows the correlation coefficients between different DDPG models and DQN model, the correlation coefficient is calculated from the vectorization of speed limits among different lanes of one round evaluated simulation. The correlation efficients between all models are very small except the one between DDPG-$r^2$ and DDPG-$r^3$. The goal of $r^3$ is to decrease the emergency braking, whereas reward $r^2$ is to increase the average speed in the bottleneck. The similarity between policies of DDPG-$r^2$ and DDPG-$r^3$ mignt be caused by the fact that the most of emergency brakings are occured in the bottleneck. Once the traffic condition at bottleneck is improved, the number of emergency braking can also be reduced. Though the DVSL policies of different DRL models are significantly different from each other, they all improved the efficiency and emission of freeway from Table. \ref{table1}. This observation confirms the need for DVSL control. Though such findings cannot be generalized to the real world freeway due to the reality gap and driver compliance, it provides useful insights for future connected autonomous vehicle highway systems.

The DVSL policies in this paper are all generated from actors consisted of neural networks. Neural
networks have long been known as ``black boxes'' because it is difficult to understand exactly how neural networks produce its output due to its large number of interacting, hidden neurons. For DVSL control problem, understanding what is learned by the neural networks is able to provide intuitions for imposing speed limits for real-world transportation systems. In this paper, we aim to understand the nerual networks for DVSL by simulating particular state inputs and observing their action--differential speed limits. Speciafically, we are particular interested in the generated speed limit variations with respect to the traffic condition of merge area. The state inputs is simulated by:
\begin{equation}
\begin{split}
s_j = [\overset{upstream}{\widetilde{0.05,0.05,0.05,0.05,0.05}}, \overset{merge}{\widehat{0.05j,0.05j,0.05j,0.05j,0.05j}}, \overset{ramp}{0.05}],\\
j = 0, 1, \cdots, 15.
\end{split}
\label{sim_state}
\end{equation}
In Eq \ref{sim_state}, the traffic in the merge area will vary from free flow to congestion when $j$ becomes larger while the traffic in the upstream section and ramp still in free flow condition. 

\begin{figure}[htbp]
  \centering
  \subfigure[The speed limit variations generated from DDPG-$r^1$]{
    \label{paa} 
    \includegraphics[width=0.45\textwidth]{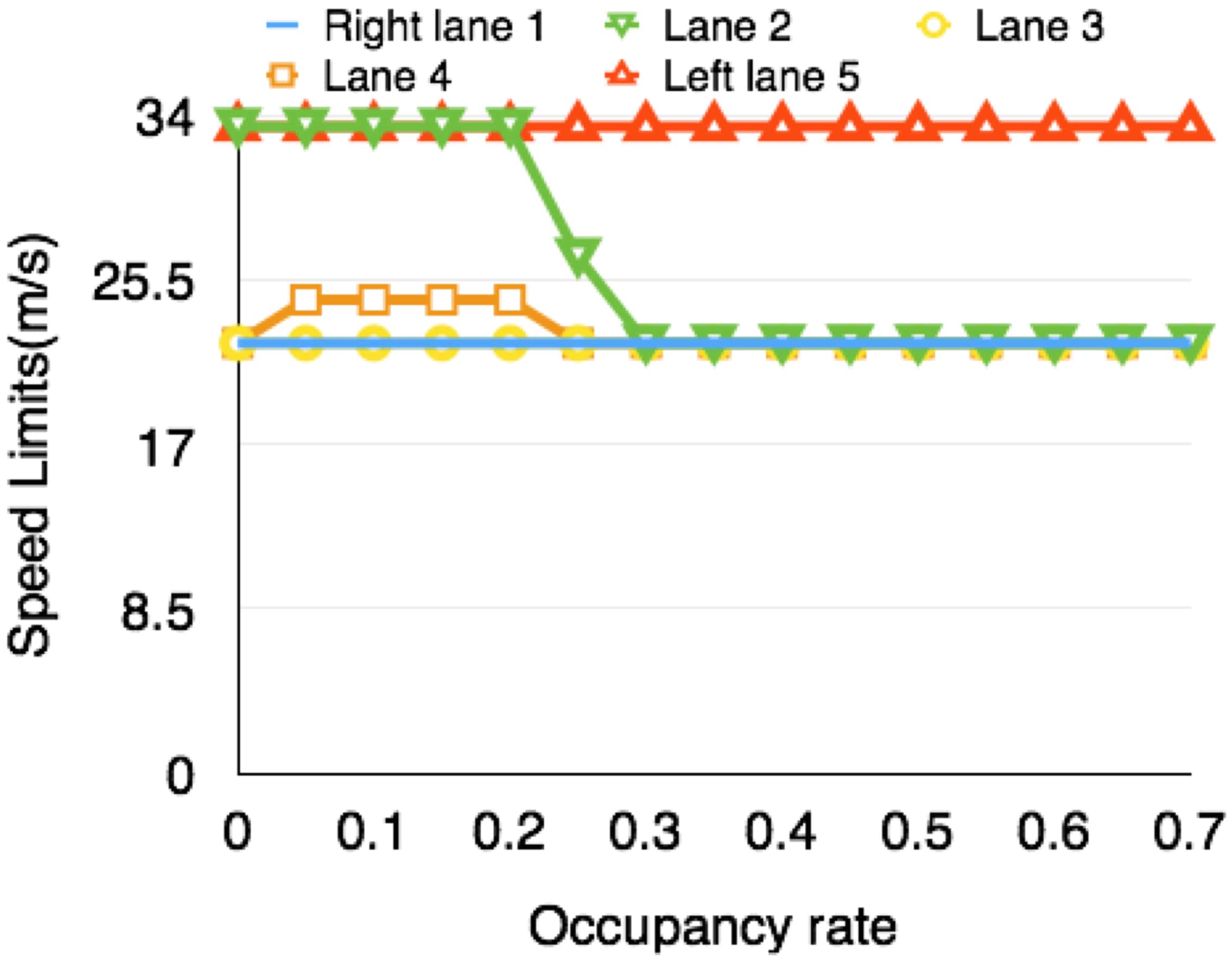}}
  \hspace{0.0in}
  \subfigure[The speed limit variations generated from DDPG-$r^2$]{
    \label{pbb} 
    \includegraphics[width=0.45\textwidth]{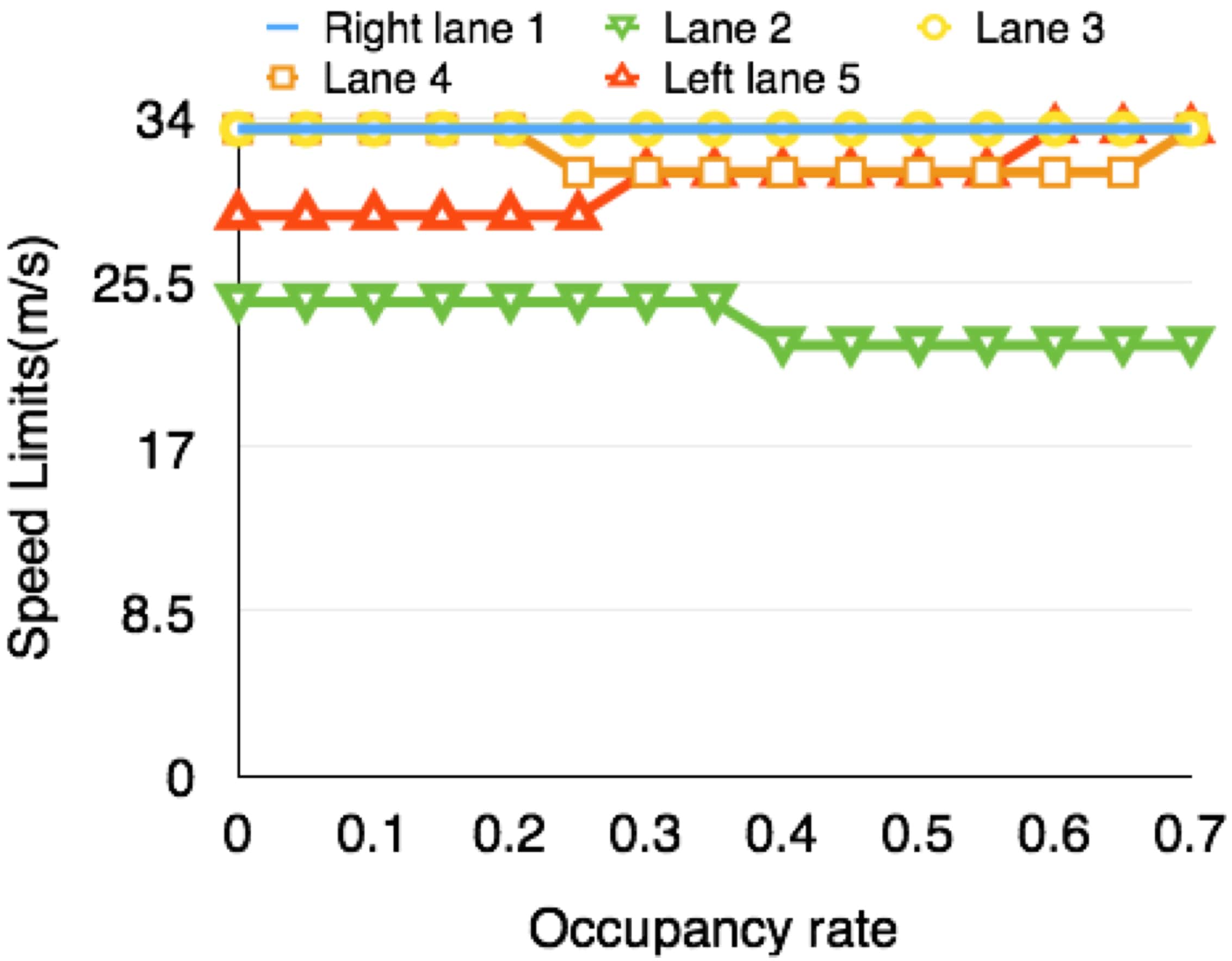}}
  \label{fig:subfig} 
  \hspace{0.0in}
    \subfigure[The speed limit variations generated from DDPG-$r^3$]{
    \label{pcc} 
    \includegraphics[width=0.45\textwidth]{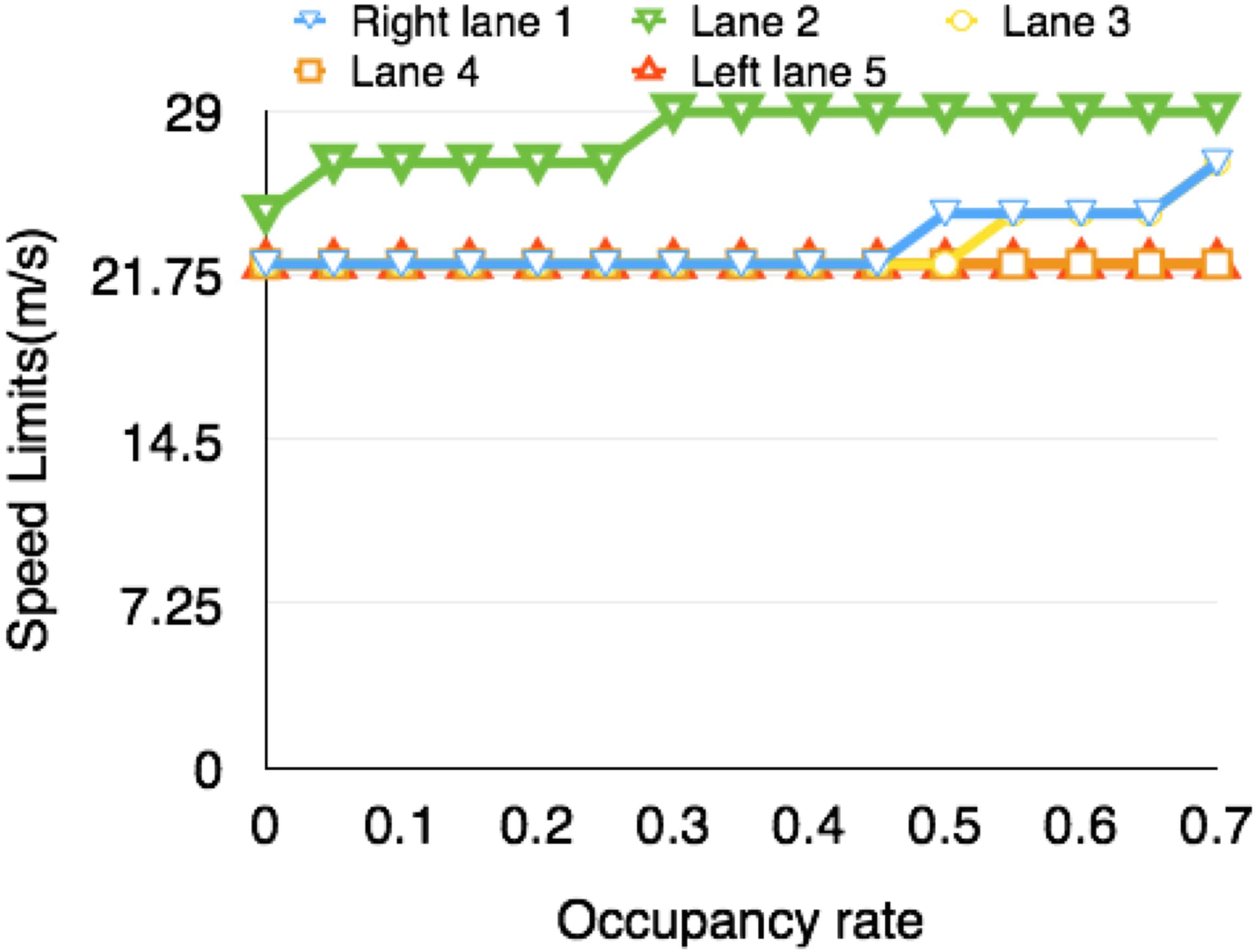}}
  \hspace{0.0in}
  \subfigure[The speed limit variations generated from DDPG-$r^4$]{
    \label{pdd} 
    \includegraphics[width=0.45\textwidth]{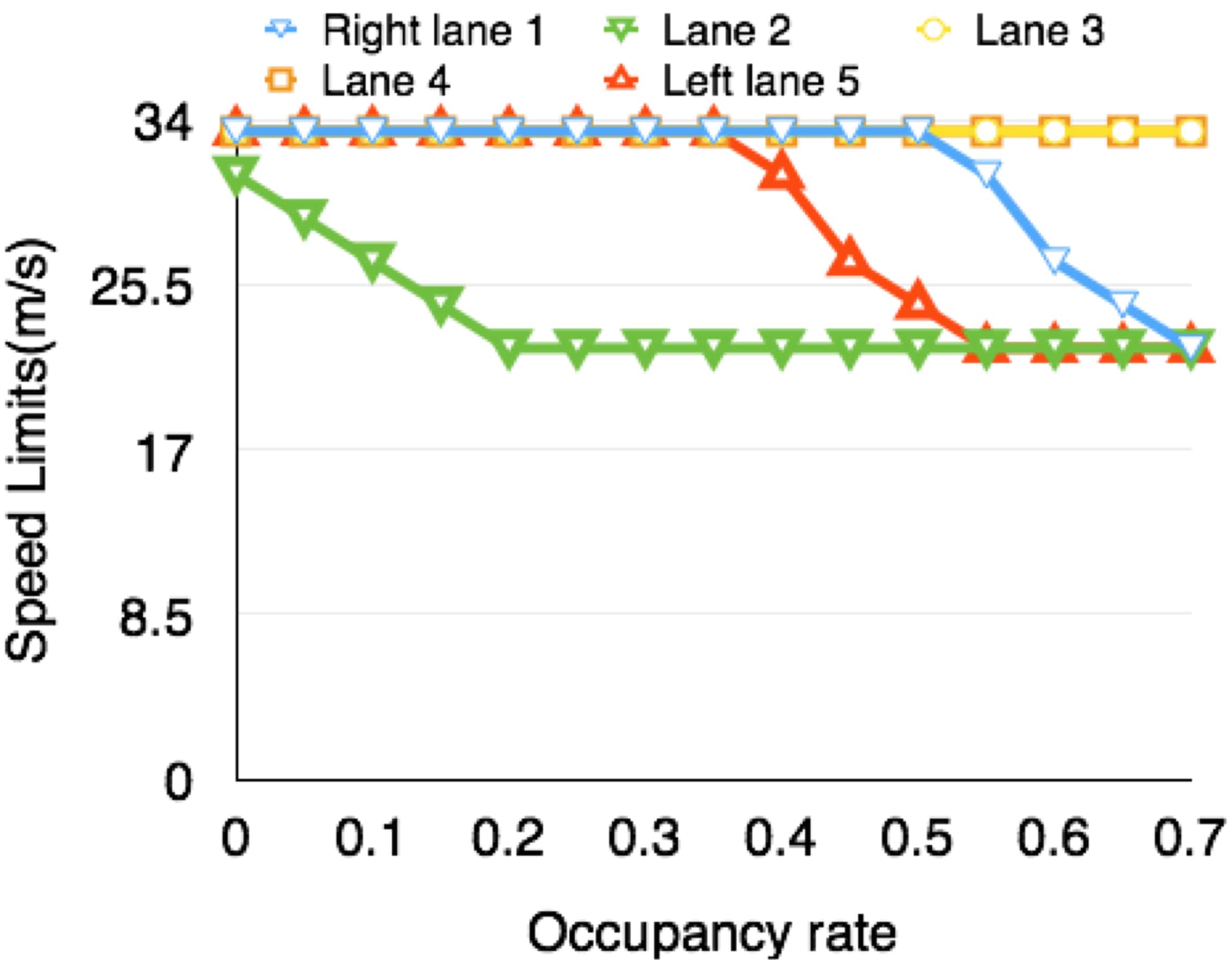}}
  \caption{Speed limit variations with respect to the occupancy rate in the merge area generated from different DDPG models}
  \label{ssma} 
\end{figure}

Fig. \ref{ssma} shows the speed limit variations with respect to the occupancy rate in the merge area. From the figures one can see that all DDPG models suggest different speed limits among lanes even the freeway is under free flow condition. The reason might be that the neural networks have learnt that the DVSL strategies can reduce the intervention between vehicles of mainstream and on-ramp, which will cause unexpected congestions, accidents and emission increases. The policies generated from DDPG-$r^1$ model are interpretable better than the policies generated from other models. No matter which condition the merge area is, the policy of DDPG-$r^1$ gives the highest velocity with 33.679m/s in the leftest lane. It indicates that the DDPG-$r^1$ assumes that the traffic in leftest lane is always in free flow condition, therefore it is not necessary to decrease the speed limit. Moreover, the velocities in lane 2 generated from DDPG-$r^1$ decreases when the occupancy rate in merge area is above 0.2. It shows that DDPG-$r^1$ has learnt that the inflow should be reduced by posing lower speed limits when the congestion has occured in the merge area. The interpretability of other DDPG models are less clear. The DDPG-$r^3$ even increases the speed limits when the congestion is occured. The reason might be that the reward signal $r^2$, $r^3$ and $r^4$ are more complex than $r^1$, thus the models trained to maximize accumulated reward $r^2$, $r^3$ and $r^4$ are more difficult to understand.

\section{Conclusion}
\label{Con}
DRL provides an efficient way to intelligently manage transportation system allowing for predictively control. To this end, a differential variable speed limits control based on deep deterministic policy gradient--a representative DRL model is proposed. Employing the actor-critc architecture of the proposed model, a large amout of discrete DVSL solutions can be efficiently learnt in a continues space. Furthermore, the experiments have shown that DDPG based DVSL control models exhibit advantages in congestion alleviation, as well as accidents and emission reduction in a high CAV simulation enviroment. In addition, the learned DVSL solution can be used to guide the VSL implementaion in real-world freeway.

Obviously, more works are required to extend the DRL based DVSL approach for larger transportation networks, where the DRL model might have to be combined with a multi-agent framework. We have studied the reward design for DRL based VSL control by comparing the performance of the models trained with different reward signals. It is noteworthy that all DRL based DVSL control models show higher evaluation indexes compared with static speed limits and variable speed limits with same speed among lanes. The results indicate a coupled structure hidden in the efficiency, safety and emission aspects of the transportation networks. The exploration of a highly abstract reward signal that can characterize all benefits will be a valuable direction in the future.

%
%
%
\appendix\label{section:references}
%
%
\bibliography{ascexmpl}

\begin{thebibliography}{}

\bibitem[\protect\citeauthoryear{}{Abdel-Aty et~al.\@}{2008}]{abdel2008dynamic}
Abdel-Aty, M., Cunningham, R., Gayah, V., and Hsia, L. (2008).
\newblock ``Dynamic variable speed limit strategies for real-time crash risk
  reduction on freeways.''\ {\em Transportation Research Record: Journal of the
  Transportation Research Board}, (2078), 108--116.

\bibitem[\protect\citeauthoryear{}{Arulkumaran
  et~al.\@}{2017}]{arulkumaran2017brief}
Arulkumaran, K., Deisenroth, M.~P., Brundage, M., and Bharath, A.~A. (2017).
\newblock ``A brief survey of deep reinforcement learning.''\ {\em arXiv
  preprint arXiv:1708.05866}.

\bibitem[\protect\citeauthoryear{}{Belletti
  et~al.\@}{2018}]{belletti2018expert}
Belletti, F., Haziza, D., Gomes, G., and Bayen, A.~M. (2018).
\newblock ``Expert level control of ramp metering based on multi-task deep
  reinforcement learning.''\ {\em IEEE Transactions on Intelligent
  Transportation Systems}, 19(4), 1198--1207.

\bibitem[\protect\citeauthoryear{}{Carlson et~al.\@}{2010}]{carlson2010optimal}
Carlson, R.~C., Papamichail, I., Papageorgiou, M., and Messmer, A. (2010).
\newblock ``Optimal mainstream traffic flow control of large-scale motorway
  networks.''\ {\em Transportation Research Part C: Emerging Technologies},
  18(2), 193--212.

\bibitem[\protect\citeauthoryear{}{Dulac-Arnold et~al.\@}{2015}]{dulac2015deep}
Dulac-Arnold, G., Evans, R., van Hasselt, H., Sunehag, P., Lillicrap, T., Hunt,
  J., Mann, T., Weber, T., Degris, T., and Coppin, B. (2015).
\newblock ``Deep reinforcement learning in large discrete action spaces.''\
  {\em arXiv preprint arXiv:1512.07679}.

\bibitem[\protect\citeauthoryear{}{Hadiuzzaman and
  Qiu}{2013}]{hadiuzzaman2013cell}
Hadiuzzaman, M. and Qiu, T.~Z. (2013).
\newblock ``Cell transmission model based variable speed limit control for
  freeways.''\ {\em Canadian Journal of Civil Engineering}, 40(1), 46--56.

\bibitem[\protect\citeauthoryear{}{Hegyi et~al.\@}{2005a}]{hegyi2005model}
Hegyi, A., De~Schutter, B., and Hellendoorn, H. (2005a).
\newblock ``Model predictive control for optimal coordination of ramp metering
  and variable speed limits.''\ {\em Transportation Research Part C: Emerging
  Technologies}, 13(3), 185--209.

\bibitem[\protect\citeauthoryear{}{Hegyi et~al.\@}{2005b}]{hegyi2005optimal}
Hegyi, A., De~Schutter, B., and Hellendoorn, J. (2005b).
\newblock ``Optimal coordination of variable speed limits to suppress shock
  waves.''\ {\em IEEE Transactions on intelligent transportation systems},
  6(1), 102--112.

\bibitem[\protect\citeauthoryear{}{Hellinga and
  Mandelzys}{2011}]{hellinga2011impact}
Hellinga, B. and Mandelzys, M. (2011).
\newblock ``Impact of driver compliance on the safety and operational impacts
  of freeway variable speed limit systems.''\ {\em Journal of Transportation
  Engineering}, 137(4), 260--268.

\bibitem[\protect\citeauthoryear{}{Kattan et~al.\@}{2015}]{kattan2015probe}
Kattan, L., Khondaker, B., Derushkina, O., and Poosarla, E. (2015).
\newblock ``A probe-based variable speed limit system.''\ {\em Journal of
  Intelligent Transportation Systems}, 19(4), 339--354.

\bibitem[\protect\citeauthoryear{}{Khondaker and
  Kattan}{2015}]{khondaker2015variable}
Khondaker, B. and Kattan, L. (2015).
\newblock ``Variable speed limit: an overview.''\ {\em Transportation Letters},
  7(5), 264--278.

\bibitem[\protect\citeauthoryear{}{Levine et~al.\@}{2018}]{levine2018learning}
Levine, S., Pastor, P., Krizhevsky, A., Ibarz, J., and Quillen, D. (2018).
\newblock ``Learning hand-eye coordination for robotic grasping with deep
  learning and large-scale data collection.''\ {\em The International Journal
  of Robotics Research}, 37(4-5), 421--436.

\bibitem[\protect\citeauthoryear{}{Li et~al.\@}{2018}]{li2018travel}
Li, L., Qu, X., Zhang, J., Li, H., and Ran, B. (2018).
\newblock ``Travel time prediction for highway network based on the ensemble
  empirical mode decomposition and random vector functional link network.''\
  {\em Applied Soft Computing}.

\bibitem[\protect\citeauthoryear{}{Li et~al.\@}{2017}]{li2017reinforcement}
Li, Z., Liu, P., Xu, C., Duan, H., and Wang, W. (2017).
\newblock ``Reinforcement learning-based variable speed limit control strategy
  to reduce traffic congestion at freeway recurrent bottlenecks.''\ {\em IEEE
  Transactions on Intelligent Transportation Systems}, 18(11), 3204--3217.

\bibitem[\protect\citeauthoryear{}{Lillicrap
  et~al.\@}{2015}]{lillicrap2015continuous}
Lillicrap, T.~P., Hunt, J.~J., Pritzel, A., Heess, N., Erez, T., Tassa, Y.,
  Silver, D., and Wierstra, D. (2015).
\newblock ``Continuous control with deep reinforcement learning.''\ {\em arXiv
  preprint arXiv:1509.02971}.

\bibitem[\protect\citeauthoryear{}{Lin et~al.\@}{2018}]{lin2018efficient}
Lin, K., Zhao, R., Xu, Z., and Zhou, J. (2018).
\newblock ``Efficient large-scale fleet management via multi-agent deep
  reinforcement learning.''\ {\em arXiv preprint arXiv:1802.06444}.

\bibitem[\protect\citeauthoryear{}{Lin}{1992}]{lin1992self}
Lin, L.-J. (1992).
\newblock ``Self-improving reactive agents based on reinforcement learning,
  planning and teaching.''\ {\em Machine learning}, 8(3-4), 293--321.

\bibitem[\protect\citeauthoryear{}{Lin et~al.\@}{2004}]{lin2004exploring}
Lin, P.-W., Kang, K.-P., and Chang, G.-L. (2004).
\newblock ``Exploring the effectiveness of variable speed limit controls on
  highway work-zone operations.''\ {\em Intelligent transportation systems},
  Vol.~8, Taylor \& Francis,  155--168.

\bibitem[\protect\citeauthoryear{}{Lu and Shladover}{2014}]{lu2014review}
Lu, X.-Y. and Shladover, S. (2014).
\newblock ``Review of variable speed limits and advisories: Theory, algorithms,
  and practice.''\ {\em Transportation Research Record: Journal of the
  Transportation Research Board}, (2423), 15--23.

\bibitem[\protect\citeauthoryear{}{Lyles et~al.\@}{2004}]{lyles2004field}
Lyles, R.~W., Taylor, W.~C., Lavansiri, D., and Grossklaus, J. (2004).
\newblock ``A field test and evaluation of variable speed limits in work
  zones.''\ {\em Transportation Research Board Annual Meeting (CD-ROM),
  Washington, DC}.

\bibitem[\protect\citeauthoryear{}{MacDonald}{2008}]{macdonald2008atm}
MacDonald, M. (2008).
\newblock ``Atm monitoring and evaluation, 4-lane variable mandatory speed
  limits 12 month report (primary and secondary indicators).''\ {\em Published
  by Department of Transport}.

\bibitem[\protect\citeauthoryear{}{Middelham}{2006}]{middelham2006dynamic}
Middelham, F. (2006).
\newblock ``Dynamic traffic management.''\ {\em Ministry of Transport, Public
  Works and Water Management, Directorate-General of Public Works and Water
  Management, AVV Transport Research Centre, Rotterdam, The Netherlands,
  Presentation to PCM Scan Team}.

\bibitem[\protect\citeauthoryear{}{Mnih et~al.\@}{2016}]{mnih2016asynchronous}
Mnih, V., Badia, A.~P., Mirza, M., Graves, A., Lillicrap, T., Harley, T.,
  Silver, D., and Kavukcuoglu, K. (2016).
\newblock ``Asynchronous methods for deep reinforcement learning.''\ {\em
  International conference on machine learning},  1928--1937.

\bibitem[\protect\citeauthoryear{}{Mnih et~al.\@}{2015}]{mnih2015human}
Mnih, V., Kavukcuoglu, K., Silver, D., Rusu, A.~A., Veness, J., Bellemare,
  M.~G., Graves, A., Riedmiller, M., Fidjeland, A.~K., Ostrovski, G., et~al.\@
  (2015).
\newblock ``Human-level control through deep reinforcement learning.''\ {\em
  Nature}, 518(7540), 529.

\bibitem[\protect\citeauthoryear{}{Ng et~al.\@}{2006}]{ng2006autonomous}
Ng, A.~Y., Coates, A., Diel, M., Ganapathi, V., Schulte, J., Tse, B., Berger,
  E., and Liang, E. (2006).
\newblock ``Autonomous inverted helicopter flight via reinforcement
  learning.''\ {\em Experimental Robotics IX}, Springer,  363--372.

\bibitem[\protect\citeauthoryear{}{Papageorgiou
  et~al.\@}{2008}]{papageorgiou2008effects}
Papageorgiou, M., Kosmatopoulos, E., and Papamichail, I. (2008).
\newblock ``Effects of variable speed limits on motorway traffic flow.''\ {\em
  Transportation Research Record: Journal of the Transportation Research
  Board}, (2047), 37--48.

\bibitem[\protect\citeauthoryear{}{Papageorgiou and
  Kotsialos}{2002}]{papageorgiou2002freeway}
Papageorgiou, M. and Kotsialos, A. (2002).
\newblock ``Freeway ramp metering: An overview.''\ {\em IEEE transactions on
  intelligent transportation systems}, 3(4), 271--281.

\bibitem[\protect\citeauthoryear{}{Piao and McDonald}{2008}]{piao2008safety}
Piao, J. and McDonald, M. (2008).
\newblock ``Safety impacts of variable speed limits-a simulation study.''\ {\em
  Intelligent Transportation Systems, 2008. ITSC 2008. 11th International IEEE
  Conference on}, IEEE,  833--837.

\bibitem[\protect\citeauthoryear{}{Roncoli et~al.\@}{2015}]{roncoli2015traffic}
Roncoli, C., Papageorgiou, M., and Papamichail, I. (2015).
\newblock ``Traffic flow optimisation in presence of vehicle automation and
  communication systems--part ii: Optimal control for multi-lane motorways.''\
  {\em Transportation Research Part C: Emerging Technologies}, 57, 260--275.

\bibitem[\protect\citeauthoryear{}{Salimans
  et~al.\@}{2017}]{salimans2017evolution}
Salimans, T., Ho, J., Chen, X., Sidor, S., and Sutskever, I. (2017).
\newblock ``Evolution strategies as a scalable alternative to reinforcement
  learning.''\ {\em arXiv preprint arXiv:1703.03864}.

\bibitem[\protect\citeauthoryear{}{Schaul
  et~al.\@}{2015}]{schaul2015prioritized}
Schaul, T., Quan, J., Antonoglou, I., and Silver, D. (2015).
\newblock ``Prioritized experience replay.''\ {\em arXiv preprint
  arXiv:1511.05952}.

\bibitem[\protect\citeauthoryear{}{Schick}{2003}]{schick2003effects}
Schick, P. (2003).
\newblock ``Effects of corridor control systems upon capacity of freeways and
  stability of traffic flow.''\ Ph.D. thesis, PhD Thesis, Institute for Road
  and Transportation Science, Faculty of Civil and Environmental Engineering,
  University of Stuttgart, PhD Thesis, Institute for Road and Transportation
  Science, Faculty of Civil and Environmental Engineering, University of
  Stuttgart.

\bibitem[\protect\citeauthoryear{}{Schulman et~al.\@}{2015}]{schulman2015trust}
Schulman, J., Levine, S., Abbeel, P., Jordan, M., and Moritz, P. (2015).
\newblock ``Trust region policy optimization.''\ {\em International Conference
  on Machine Learning},  1889--1897.

\bibitem[\protect\citeauthoryear{}{Schulman
  et~al.\@}{2017}]{schulman2017proximal}
Schulman, J., Wolski, F., Dhariwal, P., Radford, A., and Klimov, O. (2017).
\newblock ``Proximal policy optimization algorithms.''\ {\em arXiv preprint
  arXiv:1707.06347}.

\bibitem[\protect\citeauthoryear{}{Silver et~al.\@}{2016}]{silver2016mastering}
Silver, D., Huang, A., Maddison, C.~J., Guez, A., Sifre, L., Van Den~Driessche,
  G., Schrittwieser, J., Antonoglou, I., Panneershelvam, V., Lanctot, M.,
  et~al.\@ (2016).
\newblock ``Mastering the game of go with deep neural networks and tree
  search.''\ {\em nature}, 529(7587), 484.

\bibitem[\protect\citeauthoryear{}{Silver
  et~al.\@}{2014}]{silver2014deterministic}
Silver, D., Lever, G., Heess, N., Degris, T., Wierstra, D., and Riedmiller, M.
  (2014).
\newblock ``Deterministic policy gradient algorithms.''\ {\em ICML}.

\bibitem[\protect\citeauthoryear{}{Singh et~al.\@}{2002}]{singh2002optimizing}
Singh, S., Litman, D., Kearns, M., and Walker, M. (2002).
\newblock ``Optimizing dialogue management with reinforcement learning:
  Experiments with the njfun system.''\ {\em Journal of Artificial Intelligence
  Research}, 16, 105--133.

\bibitem[\protect\citeauthoryear{}{Soriguera
  et~al.\@}{2013}]{soriguera2013assessment}
Soriguera, F., Torn{\'e}, J.~M., and Rosas, D. (2013).
\newblock ``Assessment of dynamic speed limit management on metropolitan
  freeways.''\ {\em Journal of Intelligent Transportation Systems}, 17(1),
  78--90.

\bibitem[\protect\citeauthoryear{}{Sutton and
  Barto}{2017}]{sutton2017reinforcement}
Sutton, R. and Barto, A. (2017).
\newblock ``Reinforcement learning: An introduction (in preparation).

\bibitem[\protect\citeauthoryear{}{Van~der Pol and
  Oliehoek}{2016}]{van2016coordinated}
Van~der Pol, E. and Oliehoek, F.~A. (2016).
\newblock ``Coordinated deep reinforcement learners for traffic light
  control.''\ {\em Proceedings of Learning, Inference and Control of
  Multi-Agent Systems (at NIPS 2016)}.

\bibitem[\protect\citeauthoryear{}{Wei et~al.\@}{2018}]{wei2018intellilight}
Wei, H., Zheng, G., Yao, H., and Li, Z. (2018).
\newblock ``Intellilight: A reinforcement learning approach for intelligent
  traffic light control.''\ {\em Proceedings of the 24th ACM SIGKDD
  International Conference on Knowledge Discovery \& Data Mining}, ACM,
  2496--2505.

\bibitem[\protect\citeauthoryear{}{Weikl et~al.\@}{2013}]{weikl2013traffic}
Weikl, S., Bogenberger, K., and Bertini, R. (2013).
\newblock ``Traffic management effects of variable speed limit system on a
  german autobahn: Empirical assessment before and after system
  implementation.''\ {\em Transportation Research Record: Journal of the
  Transportation Research Board}, (2380), 48--60.

\bibitem[\protect\citeauthoryear{}{Wu et~al.\@}{2018}]{wu2018hybrid}
Wu, Y., Tan, H., Qin, L., Ran, B., and Jiang, Z. (2018).
\newblock ``A hybrid deep learning based traffic flow prediction method and its
  understanding.''\ {\em Transportation Research Part C: Emerging
  Technologies}, 90, 166--180.

\bibitem[\protect\citeauthoryear{}{Zhu and Ukkusuri}{2014}]{zhu2014accounting}
Zhu, F. and Ukkusuri, S.~V. (2014).
\newblock ``Accounting for dynamic speed limit control in a stochastic traffic
  environment: A reinforcement learning approach.''\ {\em Transportation
  research part C: emerging technologies}, 41, 30--47.

\bibitem[\protect\citeauthoryear{}{Zhu et~al.\@}{2017}]{zhu2017target}
Zhu, Y., Mottaghi, R., Kolve, E., Lim, J.~J., Gupta, A., Fei-Fei, L., and
  Farhadi, A. (2017).
\newblock ``Target-driven visual navigation in indoor scenes using deep
  reinforcement learning.''\ {\em Robotics and Automation (ICRA), 2017 IEEE
  International Conference on}, IEEE,  3357--3364.

\end{thebibliography}
%
%

\end{document}